%% file: ArXiv_Combined.tex
\documentclass[runningheads]{llncs}

\usepackage[T1]{fontenc} % necessary for "old text encoding" apparently (like for \k{a} (ogonek used in Polish))
\usepackage[utf8]{inputenc}

\usepackage{graphicx}
\usepackage{comment}
\usepackage{amsmath,amssymb} % define this before the line numbering.
\usepackage{color}

\usepackage[width=122mm,left=12mm,paperwidth=146mm,height=193mm,top=12mm,paperheight=217mm]{geometry}

\usepackage{epsfig}
\usepackage{graphicx}
\usepackage{amsmath}
\usepackage{amssymb}
\usepackage{xspace}

\usepackage{multirow}
\usepackage{tabularx}
\usepackage{booktabs}
\usepackage[table]{xcolor}
\usepackage{caption}

\usepackage[colorlinks=false,hyperfootnotes=false]{hyperref}
\usepackage{enumitem}
\setitemize{noitemsep,topsep=0pt,parsep=0pt,partopsep=0pt}

\input{definitions.tex}
\usepackage{pifont}% http://ctan.org/pkg/pifont

\begin{document}
% \linenumbers
\pagestyle{headings}
\mainmatter
\def\ECCVSubNumber{6440}  % Insert your submission number here

\title{Improving Optical Flow on a Pyramid Level} % Replace with your title

%\begin{comment}
%\titlerunning{ECCV-20 submission ID \ECCVSubNumber} 
%\authorrunning{ECCV-20 submission ID \ECCVSubNumber} 
%\author{Anonymous ECCV submission}
%\institute{Paper ID \ECCVSubNumber}
%\end{comment}
%******************

%\begin{comment}
\titlerunning{IOFPL - Improving Optical Flow on a Pyramid Level}
\author{Markus Hofinger$^{,\ddagger}$, Samuel Rota Bul\`o$^\dagger$, Lorenzo Porzi$^\dagger$, Arno Knapitsch$^\dagger$, Thomas Pock$^\ddagger$, Peter Kontschieder$^\dagger$
}

\institute{
	Facebook$^\dagger$, Graz University of Technology$^\ddagger$\\
	\tt\small \{markus.hofinger,pock\}@icg.tugraz.at$^\ddagger$ }
\authorrunning{M. Hofinger et al.}
%, M. Hofinger et al.}
%% First names are abbreviated in the running head.
%% If there are more than two authors, 'et al.' is used.
%%
%\institute{Princeton University, Princeton NJ 08544, USA \and
%Springer Heidelberg, Tiergartenstr. 17, 69121 Heidelberg, Germany
%%\end{comment}
%******************
\maketitle
%\unmarkedfntext{Preliminary paper version: Link to the final authenticated version to be announced}
{\let\thefootnote\relax\footnote{{Preliminary paper version: Link to the final authenticated version to be announced}}}

\input{main_matter.tex}

\appendix
%%%%%%%%% TITLE
%\twocolumn[

\clearpage
\newpage

%\unmarkedfntext{Preliminary paper version: Link to the final authenticated version to be announced}
{\let\thefootnote\relax\footnote{{Preliminary paper version: Link to the final authenticated version to be announced}}}

\begin{center}
	{\Large
		\vskip0.25cm
		\textbf{Improving Optical Flow on a Pyramid Level -- Supplementary Material}
		\vskip1cm}
\end{center}
%]
%\section{Supplementary Material}
 \input{supmat_content.tex}

\bibliographystyle{splncs04}
%\bibliography{egbib}
\bibliography{GeneralRefs}
%\bibliography{References}

\end{document}

%% file: definitions.tex
\usepackage{xspace}
\usepackage{bbold}
\usepackage{xcolor}
\usepackage{adjustbox}

\newcommand{\set}[1]{\ensuremath{\mathcal{#1}}}
\newcommand{\HDq}[1]{HD$^3$\xspace}

\newcommand{\kitti}{KITTI\xspace }

\newcommand{\epe}{EPE\xspace}
\newcommand{\pol}{Fl-all\xspace}
\newcommand{\pnoc}{Fl-noc\xspace}

\newcommand{\kitT}{Kitti 2012\xspace}
\newcommand{\kitF}{Kitti 2015\xspace}
\newcommand{\fcOne}{Flying Chairs\xspace}
\newcommand{\fc}{Flying Chairs2\xspace}
\newcommand{\ft}{Flying Things\xspace}
\newcommand{\sintel}{Sintel\xspace}

\newcommand{\Detach}{Gradient stopping\xspace}
\newcommand{\detach}{gradient stopping\xspace}

\newcommand{\tbf}[1]{{\textbf{#1}}}
\newcommand{\und}[1]{{\underline{#1}}}

\newcommand{\cmark}{\ding{51}}%
\newcommand{\xmark}{\ding{55}}%
\definecolor{mapillarygreen}{RGB}{5,203,99}

\makeatletter
\DeclareRobustCommand\onedot{\futurelet\@let@token\@onedot}
\def\@onedot{\ifx\@let@token.\else.\null\fi\xspace}

\def\eg{\emph{e.g}\onedot} \def\Eg{\emph{E.g}\onedot}
\def\ie{\emph{i.e}\onedot}

\def\wrt{w.r.t\onedot} 

\makeatother

%% file: main_matter.tex
%%%%%%%%% TITLE
\begin{center}
    \includegraphics[width=.995\textwidth]{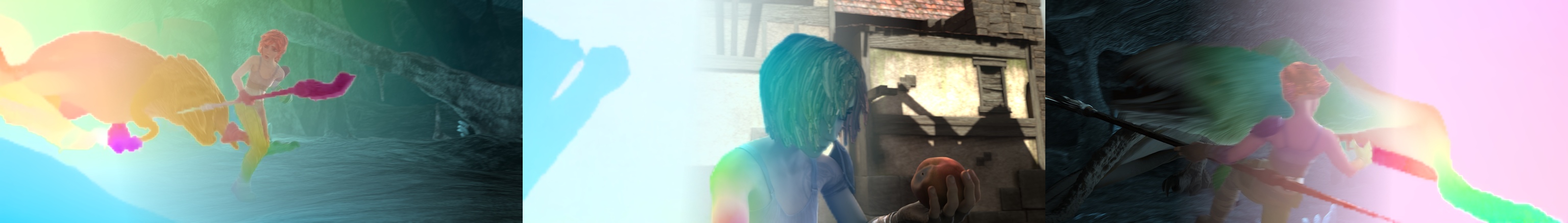}
    \includegraphics[width=.995\textwidth]{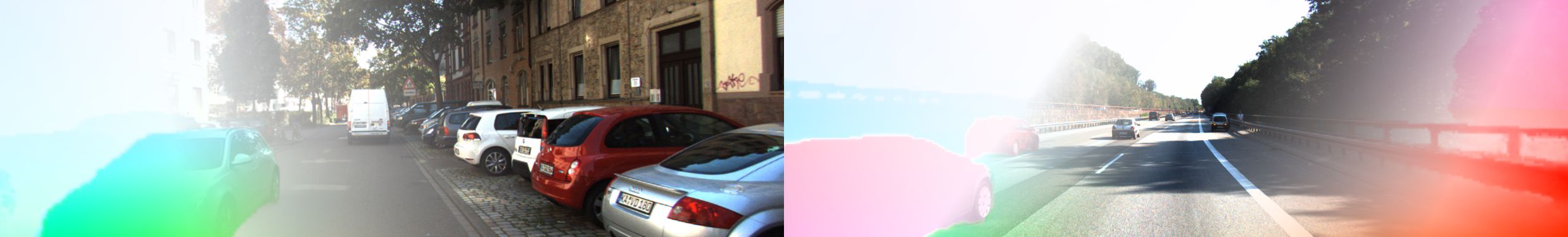}
    \captionof{figure}{Optical flow predictions from our model on images from \sintel and \kitti.}
    \label{fig:teaser}
\end{center}%

%%%%%%%%% ABSTRACT
\begin{abstract}

In this work we review the coarse-to-fine spatial feature pyramid concept, which is used in state-of-the-art optical flow estimation networks to make exploration of the pixel flow search space computationally tractable and efficient. Within an individual pyramid level, we improve the cost volume construction process by departing from a warping- to a sampling-based strategy, 
which avoids ghosting and hence enables us to better preserve fine flow details.
We further amplify the positive effects through a level-specific, loss max-pooling strategy that adaptively shifts the focus of the learning process on under-performing predictions. 
Our second contribution revises the gradient flow across pyramid levels.
The typical operations performed at each pyramid level can lead to noisy, or even contradicting gradients across levels.
We show and discuss how properly blocking some of these gradient components leads to improved convergence and ultimately better performance.
Finally, we introduce a distillation concept to counteract the issue of catastrophic forgetting during finetuning and thus preserving knowledge over models sequentially trained on multiple datasets. Our findings are conceptually simple and easy to implement, yet result in compelling improvements on relevant error measures that we demonstrate via exhaustive ablations on datasets like \fc, \ft, \sintel and \kitti. We establish new state-of-the-art results on the challenging \sintel and \kitti 2012 test datasets, and even show the portability of our findings to different optical flow and depth from stereo approaches.

\end{abstract}

%-------------------------------------------------------------------------
\section{Introduction}
State-of-the-art, deep learning based optical flow estimation methods share a number of common building blocks in their high-level, structural design. 
These blocks reflect insights gained from decades of research in \textit{classical} optical flow estimation, while exploiting the power of deep learning for further optimization of \eg performance, speed or memory constraints~\cite{Liteflownet2_hui19,PWCNetTrainingSun2018TPAMI,HD3Flow_yin2019hd3}. Pyramidal representations are among the fundamental concepts that were successfully used in optical flow and stereo matching works like~\cite{Bouguet00pyramidalimplementation}. However, while pyramidal representations enable computationally tractable exploration of the pixel flow search space, their downsides include difficulties in the handling of large motions for small objects or generating artifacts when warping occluded regions. Another observation we made is that vanilla agglomeration of hierarchical information in the pyramid is hindering the learning process and consequently leading to reduced performance. 

In this paper we identify and address shortcomings in state-of-the-art flow networks, with particular focus on improving information processing in the pyramidal representation module. For cost volume construction at a single pyramid level, we introduce a novel feature sampling strategy rather than relying on warping of high-level features to the corresponding ones in the target image. Warping is the predominant strategy in recent and top-performing flow methods~\cite{HD3Flow_yin2019hd3,Liteflownet2_hui19} but leads to degraded flow quality for fine structures. This is because fine structures require robust encoding of high-frequency information in the features, which is sometimes not recoverable after warping them towards the target image pyramid feature space. As an alternative we propose \textit{sampling}
for cost volume generation in each pyramid level, in conjunction with the sum of absolute differences as a cost volume distance function. In our sampling strategy we populate cost volume entries through distance computation between features \emph{without} prior feature warping. This helps us to better explore the complex and non-local search space of fine-grained, detailed flow transformations (see Fig.~\ref{fig:teaser}).

Using \textit{sampling} in combination with a per-pyramid level \textit{loss max-pooling} strategy further supports recovery of the motion of small and fast-moving objects. Flow errors for those objects can be attributed to the aforementioned warping issue but also because the motion of such objects often correlates with large and underrepresented flow vectors, rarely available in the training data. Loss max-pooling adaptively shifts the focus of the learning procedure towards under-performing flow predictions, without requiring additional information about the training data statistics. We introduce a loss max-pooling variant to work in hierarchical feature representations, while the underlying concept has been successfully used for dense pixel prediction tasks like semantic segmentation~\cite{RotNeuKon17cvpr}. 

Our second major contribution targets improving the gradient flow \textit{across} pyramid levels.
Functions like cost volume generation depend on bilinear interpolation, which can be shown~\cite{jiang2019linearized} to produce considerably noisy gradients.
Furthermore, fine-grained structures which are only visible at a certain pyramid level, can propagate contradicting gradients towards the coarser levels when they move in a different direction compared to their background.
Accumulating these gradients across pyramid levels ultimately inhibits convergence.
Our proposed solution is as simple as effective: by using level-specific loss terms and smartly blocking gradient propagation, we can eliminate the sources of noise.
Doing so significantly improves the learning procedure and is positively reflected in the relevant performance measures.

As minor contributions, we promote additional \textit{flow cues} that lead to a more effective generation of the cost volume. Inspired by the work of~\cite{IRR_PWC_Hur2019CVPR} that used backward warping of the optical flow to enhance the upsampling of occlusions, we advance symmetric flow networks with multiple cues (like consistencies derived from forward-backward and reverse flow information, occlusion reasoning) to better identify and correct discrepancies in the flow estimates. Finally, we also propose \textit{knowledge distillation} to counterfeit the problem of catastrophic forgetting in the context of deep-learning-based optical flow algorithms. Due to a lack of large training datasets, it is common practice to sequentially perform a number of trainings, first on synthetically generated datasets (like \fc and \ft), then fine-tuning on target datasets like \sintel or \kitti. Our distillation strategy (inspired by recent work on scene flow~\cite{SenseSceneFlow} and unsupervised approaches~\cite{SelFlow_Liu2019,DDFlow_Liu2019}) enables us to preserve knowledge from previous training steps and combine it with flow consistency checks generated from our network and further information about photometric consistency.

Our combined contributions lead to significant, cumulated error reductions over state-of-the-art networks like HD$^3$ or (variants of) PWC-Net ~\cite{HD3Flow_yin2019hd3,PWCNetTrainingSun2018TPAMI,IRR_PWC_Hur2019CVPR,BarHaim_2020_CVPR}, and we set new state-of-the-art results on the challenging \sintel and \kitti 2012 datasets. We provide exhaustive ablations and experimental evaluations on \sintel, \kitti 2012 and 2015, \ft and \fc, and significantly improve on the most important measures like \textit{Out-Noc} (percentage of erroneous non-occluded pixels) and on \textit{EPE} (average end-point-error) metrics.

%-------------------------------------------------------------------------
\section{Related Work}
\paragraph{Classical approaches.}
Optical flow has come a long way since it was introduced to the computer vision community by Lucas and Kanade~\cite{Lucas1981} and Horn and Schunck~\cite{Horn81determiningoptical}.
Following these works, the introduction of pyramidal coarse-to-fine warping frameworks were giving another huge boost in the performance of optical flow computation~\cite{Brox2004HighAO,Sun2010} -- an overview of non learning-based optical flow methods can be found in~\cite{Baker2011,Sun2014,Fortun:2015:OFM:2780699.2781074}. 

\paragraph{Deep Learning entering optical flow.} Many parts of the classical optical flow computations are well-suited for being learned by a deep neural network. Initial work using deep learning for flow was presented in~\cite{Weinzaepfel2013ICCV}, and was using a learned matching algorithm to produce semi-dense matches then refining them with a classical variational approach. The successive work of~\cite{RevaudWHS15}, whilst also relying on learned semi-dense matches, was additionally using an edge detector~\cite{Dollar2013} to interpolate dense flow fields before the variational energy minimization. End-to-end learning in a deep network for flow estimation was first done in FlowNet~\cite{DFIB15}. They use a conventional encoder-decoder architecture, and it was trained on a synthetic dataset, showing that it still generalizes well to real world datasets such as \kitti~\cite{Gei+13}. Based on this work, FlowNet2~\cite{IMSKDB17} improved by using a carefully tuned training schedule and by introducing warping into the learning framework. However, FlowNet2 could not keep up with the results of traditional variational flow approaches on the leaderboards.
SpyNet\cite{SPyNet} introduced spatial image pyramids and  PWC-Net~\cite{PWCNetSun2018,PWCNetTrainingSun2018TPAMI} additionally improved results by incorporating spatial feature pyramid processing, warping, and the use of a cost volume in the learning framework.
The flow in PWC-Net is estimated by using a stack of flattened cost volumes and image features from a Dense-Net. In~\cite{IRR_PWC_Hur2019CVPR}, PWC-Net was turned into an iterative refinement network, adding bilateral refinement of flow and occlusion in every iteration step. ScopeFlow~\cite{BarHaim_2020_CVPR} showed that improvements on top of ~\cite{IRR_PWC_Hur2019CVPR} can be achieved simply by improving training procedures. In the work of~\cite{MFF_ren2018fusion}, the group around~\cite{PWCNetSun2018} was showing further improvements on \kitF and \sintel by integrating the optical flow from an additional, previous image frame. While multi-frame optical flow methods already existed for non-learning based methods~\cite{Chaudhury1995,Werlberger2009a,Garg2013}, they were the first to show this in a deep learning framework. In~\cite{HD3Flow_yin2019hd3}, the hierarchical discrete distribution decomposition framework HD$^3$ learned probabilistic pixel correspondences for optical flow and stereo matching. It learns the decomposed match densities in an end-to-end manner at multiple scales. HD$^3$ then converts the predicted match densities into point estimates, while also producing uncertainty measures at the same time.
Devon~\cite{Lu_2020_WACV_Devon} uses a sampling and dilation based deformable cost-volume, to iteratively estimate the flow at a fixed quarter resolution in each iteration. While they showed good results on clean synthetic data, the performance on real images from \kitti was sub-optimal, indicating that sampling alone may not be sufficient.
We will show here, that integrating a direct sampling based approach into a coarse-to-fine pyramid together with LMP and Flow Cues can actually lead to very good results.
Recently, Volumetric Correspondence Networks (VCN)~\cite{VCN_NIPS2019_8367} showed that the 4D cost volume can also be efficiently filtered directly without the commonly used flattening but using separable 2D filters instead. 

\paragraph{Unsupervised methods.} Generating dense and accurate flow data for supervised training of networks is a challenging task. Thus, most large-scale datasets are synthetic~\cite{Butler_ECC_2012,DFIB15,FlowNet3_Ilg2018}, and real data sets remained small and sparsely labeled \cite{Menze2018JPRS,Menze2015ISA}.
Unsupervised methods do not rely on that data, instead, those methods usually utilize the photometric loss between the original image in the warped, second image to guide the learning process~\cite{UnsupFlow2016jjyu}. However, the photometric loss does not work for occluded image regions, and therefore methods have been proposed to generate occlusion masks beforehand or simultaneously~\cite{Aodha2013_LCMFOF,Yamaguchi14}. 

\paragraph{Distillation.} To learn the flow values of occluded areas, DDFlow~\cite{DDFlow_Liu2019} is using a student-teacher network which distills data from reliable predictions, and uses these predictions as annotations to guide a student network. SelFlow~\cite{SelFlow_Liu2019} is built in a similar fashion but vastly improves the quality of the flow predictions in occluded areas by introducing a superpixel-based occlusion hallucination technique. They obtain state-of-the-art results when fine-tuning on annotated data after pre-training in a self-supervised setting. 
SENSE~\cite{SenseSceneFlow} tries to integrate optical flow, stereo, occlusion, and semantic segmentation in one semi-supervised setting. Much like in a multi-task learning setup, SENSE~\cite{SenseSceneFlow} uses a shared encoder for all four tasks, which can exploit interactions between the different tasks and leads to a compact network. SENSE uses pre-trained models to “supervise” the network on data with missing ground truth annotations using a distillation loss~\cite{Hin+15}. To couple the four tasks, a self-supervision loss term is used, which largely improves regions without ground truth (\eg sky regions). 

%-------------------------------------------------------------------------
\section{Main Contributions}
\label{sec:main}

In this section we review pyramid flow network architectures~\cite{PWCNetSun2018,HD3Flow_yin2019hd3}, and propose a set of modifications to the pyramid levels (\S~\ref{sec:pyramid-level}) and their training strategy (\S~\ref{sec:training}), which work in a synergistic manner to greatly boost performance.

\subsection{Pyramid flow networks}
Pyramid flow networks (PFN) operate on pairs of images, building feature pyramids with decreasing spatial resolution using ``siamese'' network branches with shared parameters.
Flow is iteratively refined starting from the top of the pyramid, each layer predicting an offset relative to the flow estimated at the previous level.
For more details about the operations carried out at each level see \S~\ref{sec:pyramid-level}.

\paragraph{Notation.}
We represent multi-dimensional feature maps as functions $I_i^l:\set{I}^l_i\to\mathbb{R}^d$, where $i=1,2$ indicates which image the features are computed from, $l$ is their pyramid level, and $\set{I}^l_i\subset\mathbb{R}^2$ is the set of pixels of image $i$ at resolution $l$.
We call \emph{forward flow} at level $l$ a mapping $F^l_{1\to 2}:\set I^l_1\to\mathbb R^2$, which intuitively indicates where pixels in $I^l_1$ moved to in $I^l_2$ (in relative terms).
We call \emph{backward flow} the mapping $F^l_{2\to 1}:\set I^l_2\to\mathbb R^2$ that indicates the opposite displacements.
Pixel coordinates are indexed by $u$ and $v$, \ie $x=(x_u, x_v)$, and given $x\in\set I^l_1$, we assume that $I^l_1(x)$ implicitly applies bilinear interpolation to read values from $I^l_1$ at sub-pixel locations.

\subsection{Improving pyramid levels in PFNs}
\label{sec:pyramid-level}
Many PFNs~\cite{PWCNetSun2018,HD3Flow_yin2019hd3} share the same high-level structure in each of their levels. %, depicted in Fig.~\ref{fig:network_structure}. 
First, feature maps from the two images are aligned using the coarse flow estimated in the previous level, and compared by some distance function to build a cost volume (possibly both in the \emph{forward} and \emph{backward} directions).
Then, the cost volume is combined with additional information from the feature maps (and optionally additional ``flow cues'') and fed to a ``decoder'' subnet.
This subnet finally outputs a residual flow, or a match density from which the residual flow can be computed.
A separate loss is applied to each pyramid layer, providing deep supervision to the flow refinement process.
In the rest of this section, we describe a set of generic improvements that can be applied to the pyramid layers of several state of the art pyramid flow networks.

\subsubsection{Cost volume construction}
\label{sec:cost-volume}
The first operation at each level of most pyramid flow networks involves comparing features between $I^l_1$ and $I^l_2$, conditioned on the flow $F^{l-1}_{1\to 2}$ predicted at the previous level.
In the most common implementation, $I^l_2$ is warped using $F^{l-1}_{1\to 2}$, and the result is cross-correlated with $I^l_1$.
More formally, given $I^l_2$ and $F^{l-1}_{1\to 2}$, the warped image is given by $I^l_{2\to 1}(x)=I^l_2(x+F^{l-1}_{1\to 2}(x))$ and the cross-correlation is computed with:
\begin{equation}
\label{eq:v_warp}
V^\text{warp}_{1\to 2}(x,\delta)=I^l_1(x)\cdot I^l_{2\to 1}(x+\delta)
=I^l_1(x)\cdot I^l_2(x+\delta+F^{l - 1}_{1\to 2}(x+\delta))\,,
\end{equation}
where $\delta\in[-\Delta,\Delta]^2$ is a restricted search space and $\cdot$ is the vector dot product.
This warping operation, however, suffers from a serious drawback which occurs when small regions move differently compared to their surroundings.

\begin{figure}[t]
	\centering
	
	\includegraphics[width=0.55\columnwidth]{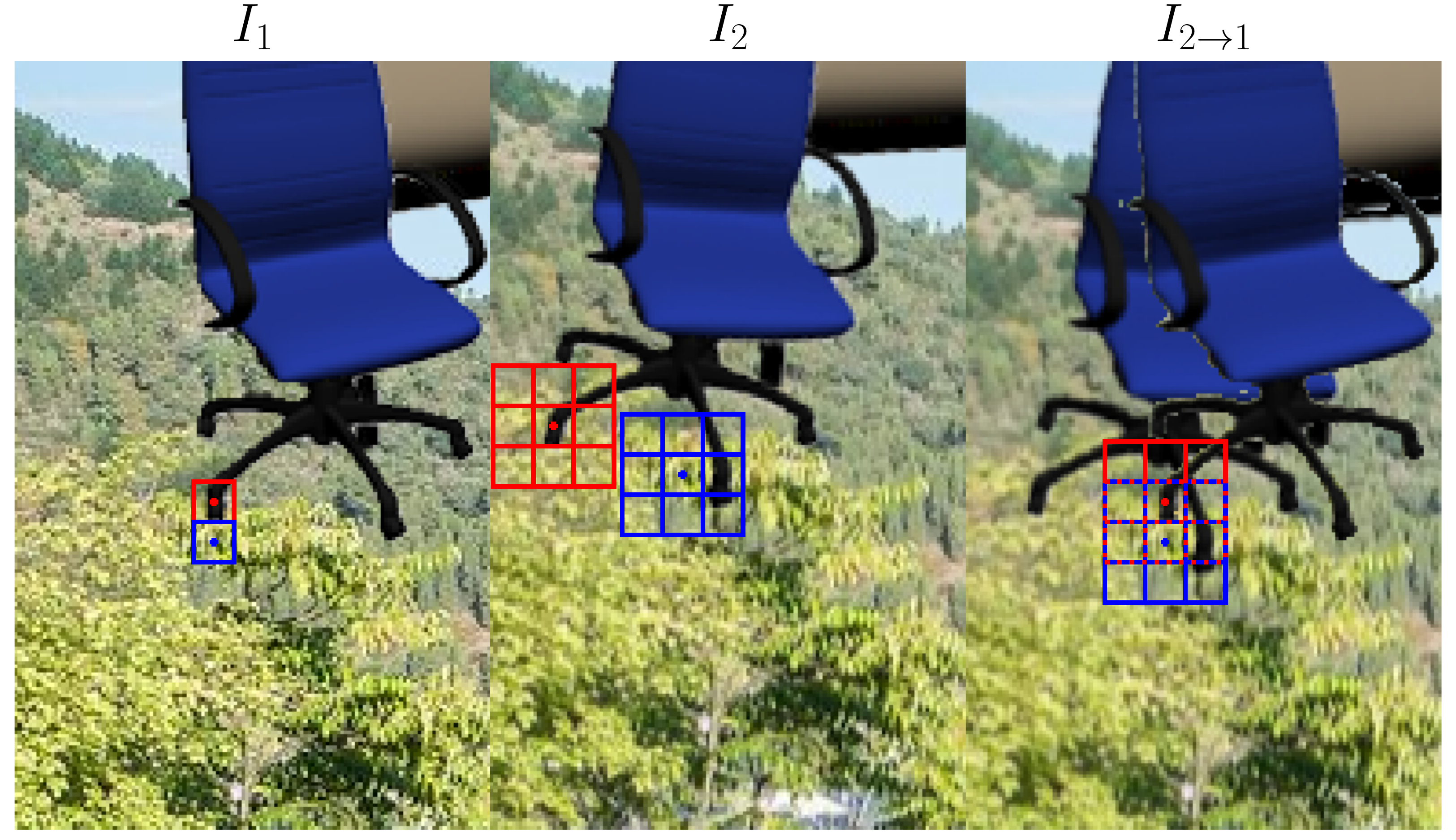}
	\includegraphics[width=0.44\columnwidth]{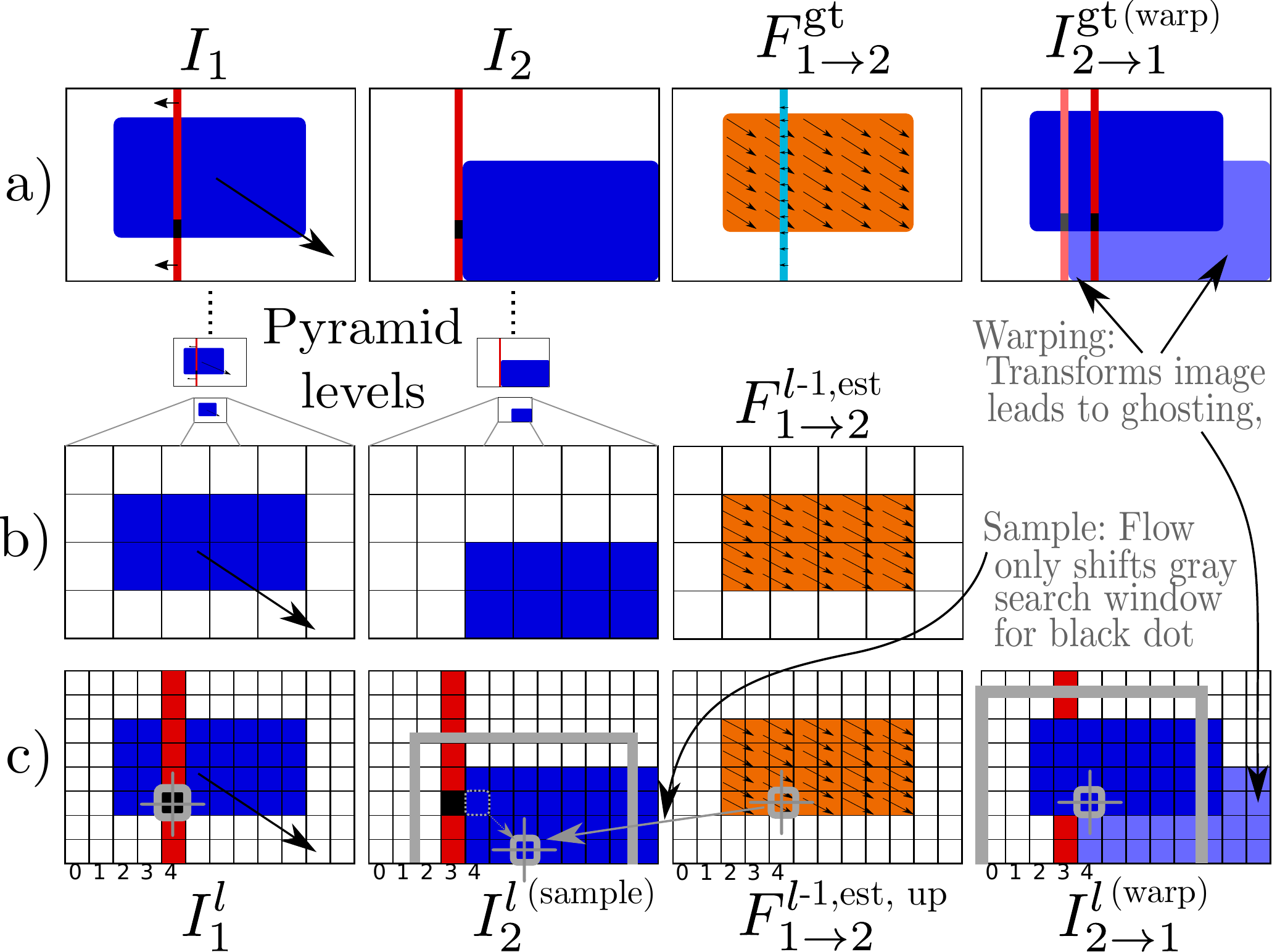}
	\caption{
		Sampling vs. Warping.
		\textbf{Left}:
		Warping leads to image ghosting in the warped image $I_{2\to1}$;
		Also, neighbouring pixels in $I_1$ must share parts of their search windows in $I_{2\to1}$, while for sampling they are independently sampled from the original image $I_2$.
		\textbf{Right}: A toy example;
		a) Two moving objects: a red line with a black dot and a blue box. Warping with $F_{1\to2}^{\text{gt}}$ leads to ghosting effects.
		b) Zooming into lowest pyramid resolution shows loss of small details due to down-scaling.
		c) \emph{Warping} $I_2^l$ with the flow estimate from the coarser level leads to distortions in $I_{2\to1}^l$ (the black dot gets covered up).		Instead, direct \emph{sampling} in $I_2^l$ with a search window(gray box) that is  offset by the flow estimate avoids these distortions and hence leads to more stable correlations.
		\vspace{-0.5cm}}
	\label{fig:SamplingCorr_vsWarpCorr}
\end{figure}

\begin{figure}[t]
    \centering
    \includegraphics[width=\columnwidth]{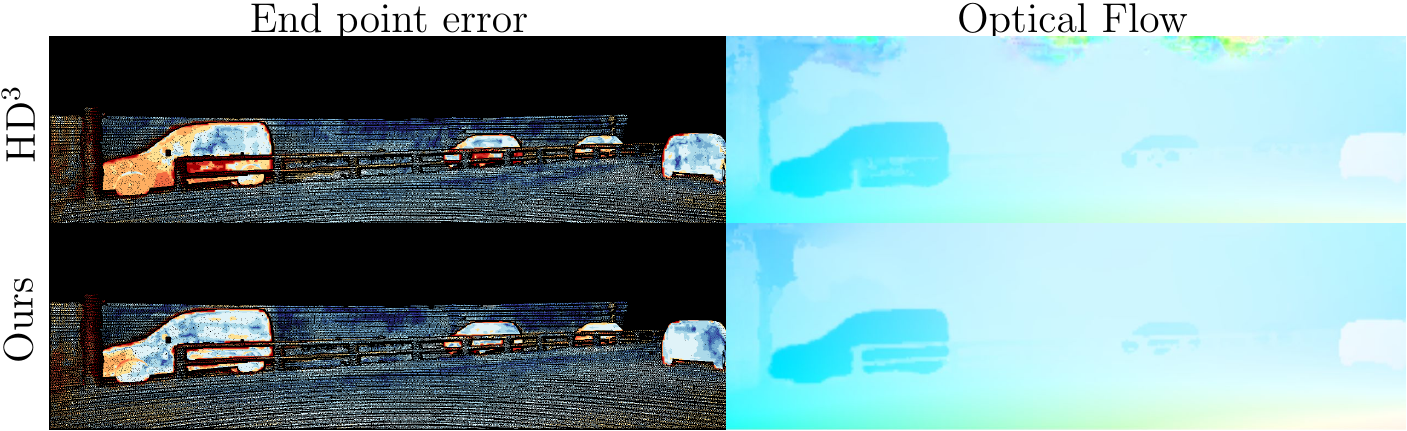}
    \caption{Predicted optical flow and end point error on \kitti obtained with HD$^3$ from the model zoo (top) and our IOFPL version (bottom). Note how our model is better able to preserve small details.}
    \label{fig:vis_comparison_ours_vs_theirs}
\end{figure}

This case is represented in Fig.~\ref{fig:SamplingCorr_vsWarpCorr}: A small object indicated by a red line moves in a different direction than a larger blue box in the background.
As warping uses the coarse flow estimate from the previous level, which cannot capture fine-grained motions, there is a chance that the smaller object gets lost during the feature warping. This makes it undetectable in $I_{2\to1}^l$, even with an infinite 
cost volume range (CVr/CV-range) $\delta$.
%search range $\Delta$.
To overcome this limitation, we propose a different cost volume construction strategy, which exploits direct sampling operations.
This approach always accesses the original, undeformed features $I^l_2$, without loss of information, and the cross-correlation in Eq.~\eqref{eq:v_warp} now becomes:
\begin{equation} 
\label{eq:v_samp_corr}
V^\text{samp,Corr}_{1\to 2}(x,\delta)=I^l_1(x)\cdot I^l_2(x+\delta+F^{l-1}_{1\to 2}(x))\,.
\end{equation}
For this operator, the flow just acts as an offset that sets the center of the correlation window in the feature image $I_2^l$.
Going back to Fig.~\ref{fig:SamplingCorr_vsWarpCorr}, one can see that the sampling operator is still able to detect the small object, as it is also exemplified on real data in Fig.~\ref{fig:vis_comparison_ours_vs_theirs}.
In contrast to \cite{Lu_2020_WACV_Devon}, our approach still uses the coarse to fine pyramid and hence doesn't require dilation in the cost volume for large motions.
In our experiments we also consider a variant where the features are compared in terms of Sum of Absolute Differences (SAD) instead of a dot product:
\begin{equation}
\label{eq:v_samp_sad}
V^\text{samp,SAD}_{1\to 2}(x,\delta)=\|I^l_1(x) - I^l_2(x+\delta+F^{l-1}_{1\to 2}(x))\|_1\,.
\end{equation}

\subsubsection{Loss Max Pooling}
We apply a Loss Max-Pooling (LMP) strategy~\cite{RotNeuKon17cvpr}, also known as Online Hard Example Mining (OHEM), to our knowledge for the first time in the context of optical flow.
In our experiments, and consistent with the findings in~\cite{RotNeuKon17cvpr}, we observe that LMP can help to better preserve small details in the flow.
The total loss is the sum of a pixelwise loss $\ell_x$ over all $x\in\set I_1$, but we optimize a weighted version thereof that selects a fixed percentage of the highest per-pixel losses.
The percentage value $\alpha$ is best chosen according to the quality of the ground-truth in the target dataset. 
This can be written in terms of a loss max-pooling strategy as follows:
\begin{equation}
L=\max\left\{\sum_{x\in\set I_1}w_x\ell_x\,:\, \Vert w\Vert_1\leq 1\,, \Vert w\Vert_\infty\leq\frac{1}{\alpha |\set I_1|}\right\}\,,
\end{equation}
which is equivalent to putting constant weight $w_x=\frac{1}{\alpha |\set I_1|}$ on the percentage of pixels $x$ exhibiting the highest losses, and setting $w_x=0$ elsewhere. 

LMP lets the network focus on the more difficult areas of the image, while reducing the amount of gradient signals where predictions are already correct.
To avoid focussing on outliers, we set the loss to 0 for pixels that are out of reach for the current relative search range $\Delta$.
For datasets with sparsely annotated ground-truth, like \eg \kitti~\cite{Gei+13}, we re-scale the per pixel losses $\ell_x$ to reflect the number of valid pixels.
Note that, when performing distillation, loss max-pooling is only applied to the supervised loss, in order to further reduce the effect of noise that survived the filtering process described in \S~\ref{sec:distillation}.

\subsection{Improving gradient flow across PFN levels}
\label{sec:training}

Our quantitatively most impacting contribution relates to the way we pass gradient information across the different levels of a PFN.
In particular, we focus on the bilinear interpolation operations that we implicitly perform on $I^l_2$ while computing Eq.s~\eqref{eq:v_warp}, \eqref{eq:v_samp_corr} and \eqref{eq:v_samp_sad}.
It has been observed~\cite{jiang2019linearized} that taking the gradient of bilinear interpolation w.r.t. the sampling coordinates (\ie the flow $F^{l-1}_{1\to 2}$ from the previous level in our case) is often problematic.
To illustrate the reason, we restrict our attention to the 1-D case for ease of notation, and write linear interpolation from a function $\hat{f}:\mathbb{Z}\to\mathbb{R}$:
\begin{equation}
  f(x) = \sum_{\eta\in\{0,1\}} \hat{f}(\lfloor x \rfloor + \eta) \left[(1-\eta) (1 - \tilde{x}) + \eta \tilde{x} \right]\,,
\end{equation}
where $\tilde{x}=x-\lfloor x \rfloor$ denotes the fractional part of $x$.
The derivative of the interpolated function $f(x)$ with respect to $x$ is:
\begin{equation}
  \frac{df}{dx}(x) = \sum_{\eta\in\{0,1\}} \hat{f}(\lfloor x \rfloor + \eta) (2\eta-1)\,.
\end{equation}
The gradient function $\frac{df}{dx}$ is discontinuous, for its value drastically changes as $\lfloor x\rfloor$ crosses over from one integer value to the next, possibly inducing strong noise in the gradients.
An additional effect, specific to our case, is related to the issues already highlighted in \S~\ref{sec:cost-volume}: since $F^{l-1}_{1\to 2}$ is predicted at a lower resolution than level $l$ operates at, it cannot fully capture the motion of smaller objects.
When this motion contrasts with that of the background, the gradient w.r.t. $F^{l-1}_{1\to 2}$ produced from the sampling at level $l$ will inevitably disagree with that produced by the loss at level $l-1$, possibly slowing down convergence.

\begin{figure}[b]
	\centering
	\includegraphics[width=0.95\columnwidth]{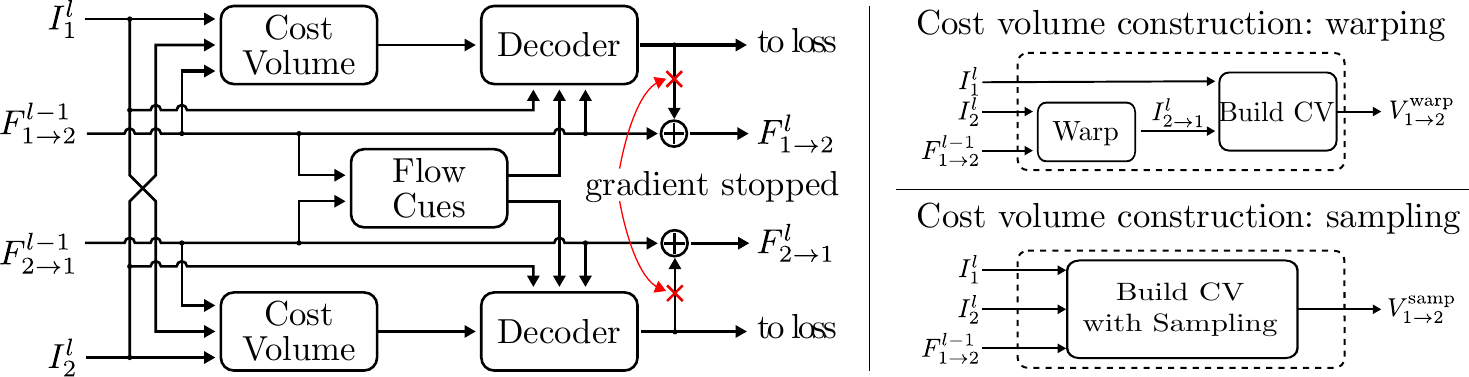}
	\caption{Left: Network structure -- flow estimation per pyramid level; Gradients are stopped (red cross); Right: Cost volume computation with sampling vs. warping.}
	\label{fig:network_structure}
\end{figure}

While~\cite{jiang2019linearized} proposes a different sampling strategy to reduce the noise issues discussed above, in our case we opt for a much simpler work around.
Given the observations about layer disagreement, and the fact that the loss at $l - 1$ already provides direct supervision on $F^{l-1}_{1\to 2}$, we choose to stop back-propagation of partial flow gradients coming from higher levels, as illustrated in Fig.~\ref{fig:network_structure}.

\begin{figure}[t]
	\centering
	\includegraphics[width=0.99\textwidth]{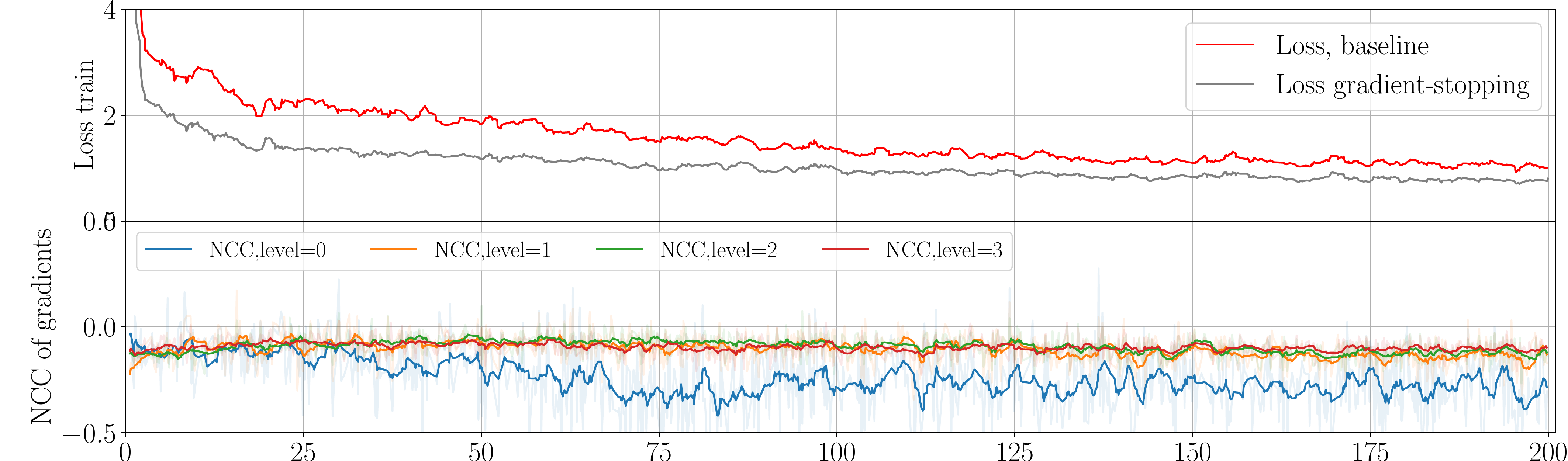}
	\caption{
		Top: Loss of model decreases when the flow gradient is stopped;
		Bottom: Partial gradients coming from the current level loss and the next level via the flow show a negative Normalized Cross Correlation (NCC), indicating that they oppose each other.
	}
	\label{fig:cross_corr}
\end{figure}

Evidence for this effect can be seen in Fig.~\ref{fig:cross_corr}, where the top shows the development of the training loss for a Flying Chairs 2 training with an HD$^3$ model.
The training convergence clearly improves when the partial flow gradient is stopped between the levels (red cross in  Fig.~\ref{fig:network_structure}).
On the bottom of the figure the Normalized Cross Correlation (NCC) between the partial gradient coming from the next level via the flow and the current levels loss is shown.
On average the correlation is negative, indicating that for each level of the network the partial gradient that we decided to stop (red cross), coming from upper levels, points in a direction that opposes the partial gradient from the loss directly supervising the current level, thus harming convergence.
Additional evidence of the practical, positive impact of our gradient stopping strategy is given in the experiment section \S~\ref{ssec:flowAb}.

Further evidence on this issue can be gained by analyzing the parameters gradient variance \cite{Wang_NIPS2013_5034_VarianceSGD} as it impacts the rate of convergence for stochastic gradient descent methods. 
Also the $\beta$-smoothness \cite{Nips2018_Santurkar_Batchnorm} of the loss function gradient can give similar insights.
In the supplementary material (section \S~A) we provide further experiments that show that gradient stopping also helps to improve these properties, and works for stereo estimation and other flow models as well.

\subsection{Additional refinements}\label{ssec:AdditionalRefinements}

\subsubsection{Flow cues}
As mentioned at the beginning of \S~\ref{sec:pyramid-level}, the decoder subnet in each pyramid level processes the raw feature correlations to a final cost volume or direct flow predictions.
To provide the decoder with contextual information, it commonly~\cite{PWCNetSun2018,HD3Flow_yin2019hd3} also receives raw features (\ie $I^l_1$, $I^l_2$ for forward and backward flow, respectively).
Some works~\cite{OccAwareUnsupLearning__Wang2018_CVPR,IRR_PWC_Hur2019CVPR,FlowNet3_Ilg2018} 
also append other cues, in the form of hand-crafted features, aimed at capturing additional prior knowledge about flow consistency.
Such flow cues are cheap to compute but otherwise hard to learn for CNNs as they require various forms of non-local spatial transformations.
In this work, we propose a set of such flow cues that provides mutual beneficial information, and perform very well in practice when combined with costvolume sample and LMP (see \S~\ref{ssec:flowAb}).
These cues are namely forward-backward flow warping, reverse flow estimation, map uniqueness density and out-of-image occlusions, and are described in detail in the supplementary material (\S~B).

\subsubsection{Knowledge distillation}
\label{sec:distillation}
\begin{figure}[t]
    \centering
     \includegraphics[width=0.95\textwidth]{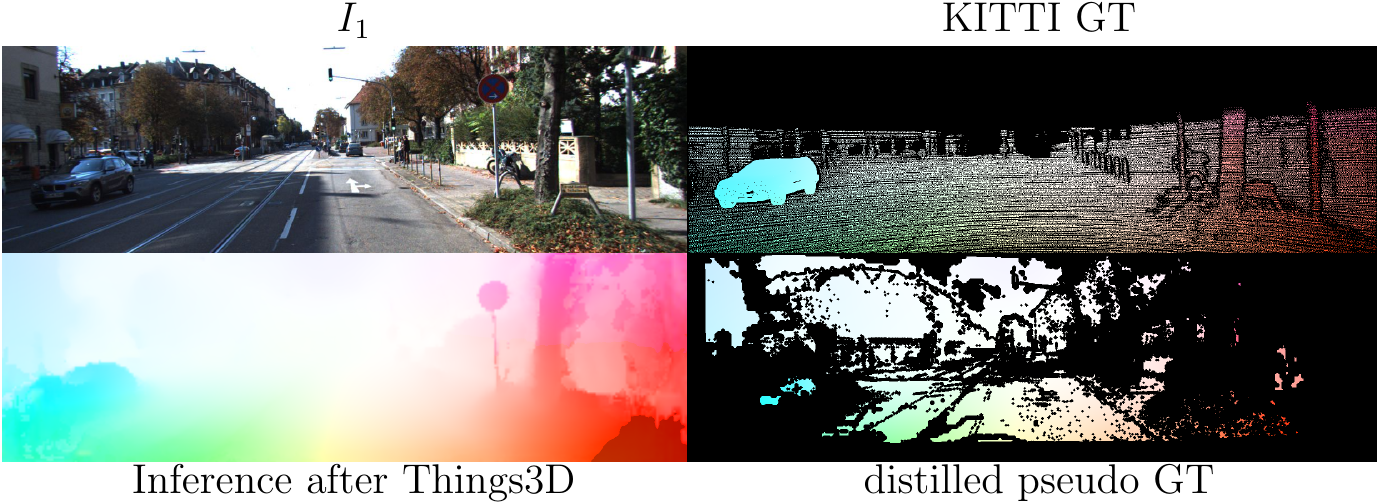}
    \caption{Illustration of our data distillation process. Left to right: input image and associated \kitti ground truth, dense prediction from a Flying Things3D-trained network and pseudo-ground truth derived from it.}
    \label{fig:Data_distillation_Concept}
\end{figure}

\begin{figure}[t]
    \centering
    \includegraphics[width=\columnwidth]{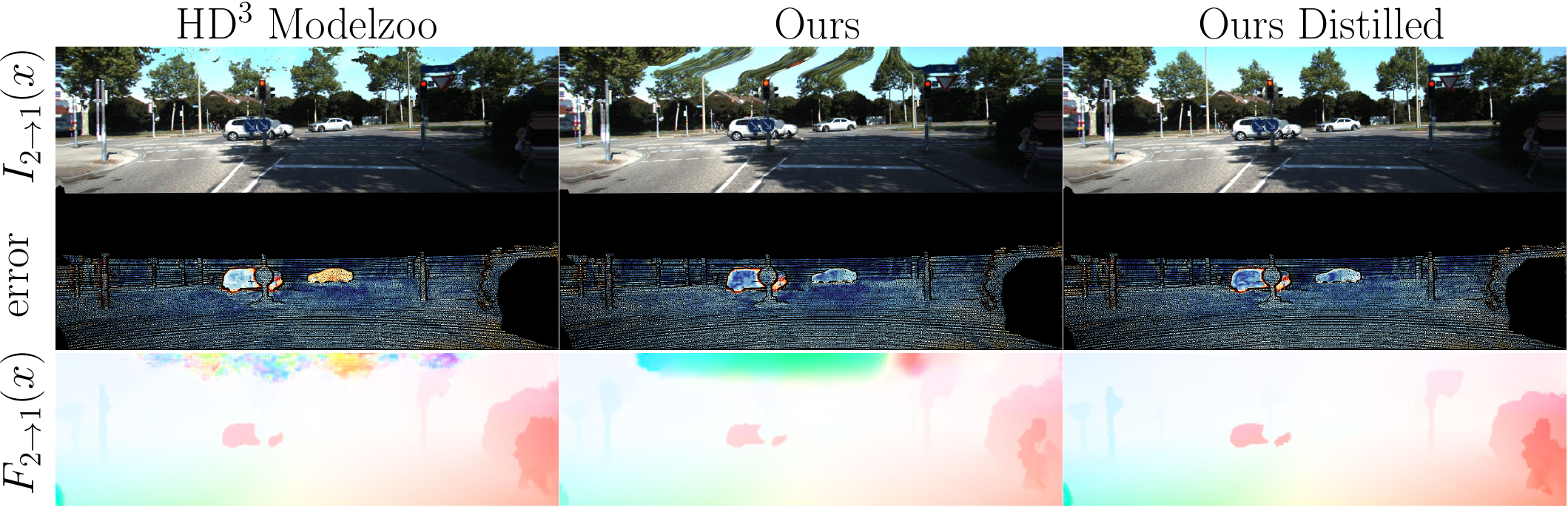}
    \caption{Qualitative results on \kitti for:  HD$^3$ modelzoo (left), our version with all contributions except distillation (center), and with distillation (right).}
    \label{fig:vis_comparison_distilled_nondistilled}
\end{figure}

Knowledge distillation~\cite{Hin+15} consists in extrapolating a training signal directly from another trained network, ensemble of networks, or perturbed networks~\cite{RotPorKon16}, typically by mimicking their predictions on some available data.
In PFNs, distillation can help to overcome issues such as lack of flow annotations on \eg sky, which results in cluttered outputs in those areas.
Formally, our goal is to distill knowledge from a pre-trained master network (\eg on \fc and/or \ft) by augmenting a student network with an additional loss term, which tries to mimic the predictions the master produces on the input at hand (Fig.~\ref{fig:Data_distillation_Concept}, bottom left).
At the same time, the student is also trained with a standard, supervised loss on the available ground-truth (Fig.~\ref{fig:Data_distillation_Concept}, top right).
In order to ensure a proper cooperation between the two terms, we prevent the distillation loss from operating blindly, instead enabling it selectively based on a number of consistency and confidence checks (refer to the supplementary material for details).
Like for the ground-truth loss, the data distillation loss is scaled with respect to the valid pixels present in the pseudo ground-truth.
The supervised and the distillation losses are combined into a total loss 
\begin{align}
\mathcal{L} =  \alpha \mathcal{L}_S +  (1 - \alpha) \mathcal{L}_D 
\end{align} 
with the scaling factor $\alpha =0.9$.
A qualitative representation of the effects of our proposed distillation on \kitti data is given in Fig.~\ref{fig:vis_comparison_distilled_nondistilled}.

%-------------------------------------------------------------------------

%------------------------------------------------------------------------
\section{Experiments}
We assess the quality of our contributions by providing a number of exhaustive ablations on \fcOne, \fc, \ft, \sintel, \kitT and \kitF.
We ran the bulk of ablations based on HD$^3$~\cite{HD3Flow_yin2019hd3}, \ie a state-of-the-art, 2-frame optical flow approach.
We build on top of their publicly available code 
and stick to default configuration parameters where possible, and describe and re-train the baseline model when deviating.

The remainder of this section is organized as follows. We provide i) in \S~\ref{ssec:Setup} a summary about the experimental and training setups and our basic modifications over HD$^3$, ii) in \S~\ref{ssec:flowAb} an exhaustive number of ablation results for all aforementioned datasets by learning \textbf{only} on the \fc training set, and for all reasonable combinations of our contributions described in  \S~\ref{sec:main}, as well as ablations on \sintel, and iii) list and discuss in \S~\ref{ssec:benchmarks} our results obtained on the \kitT, \kitF and \sintel test datasets, respectively.
In the supplementary material we further provide 
i) more technical details and ablation studies about the used \textit{flow cues}, 
%ii) results when applying \detach to other optical flow networks and depth from stereo, 
%ii) results on gradient stopping, like smoothness and variance analysis or with a PWC baseline or depth from stereo
ii) smoothness and variance analyses for gradient stopping and its impact on depth from stereo or with a PWC baseline
iii) ablations on extended search ranges for the cost volume, and 
iv) ablations on distillation.

\subsection{Setup and modifications over HD$^3$}
\label{ssec:Setup}
We always train on 4xV100 GPUs with 32GB RAM using PyTorch, and obtain additional memory during training by switching to In-Place Activated BatchNorm (non-synchronized, Leaky-ReLU)~\cite{RotPorKon18a}.
We decided to train on \fc rather than \fcOne for our main ablation experiments, since it provides ground truth for both, forward and backward flow directions. Other modifications are experiment-specific and described in the respective sections.

\paragraph{Flow - Synthetic data pre-training.}
Also the \ft dataset provides ground truth flow for both directions. We always train and evaluate on both flow directions, since this improves generalization to other datasets. We use a batch size of 64 to decrease training times and leave the rest of configuration parameters unchanged \wrt the default HD$^3$ code. 

\paragraph{Flow - Fine-tuning on \kitti.}
Since both the \kitT and the \kitF datasets are very small and only provide forward flow ground truth, we follow the \HDq{}  training protocol and join all \kitti training sequences for the final fine-tuning (after pre-training on \fc and \ft). However, we ran independent multi-fold cross validations and noticed faster convergence of our model over the baseline. We therefore perform early stopping after 1.6k (CVr$\pm4$)/ 1.4k (CVr$\pm8$) epochs, to prevent over-fitting. Furthermore, before starting the fine-tuning process of the pre-trained model, we label the \kitti training data for usage described in the knowledge distillation paragraph in \S~\ref{ssec:AdditionalRefinements}.

\paragraph{Flow - Fine-tuning on \sintel.}
We only train on all the images in the \emph{final} pass and ignore the \emph{clean} images like HD$^3$ for comparability. Also, we only use the forward flow ground truth since backward flow ground truth is unavailable. Although not favorable, our model can still be trained in this setting since we use a single, shared set of parameters for the forward and the backward flow paths. 
We kept the original 1.2k finetuning iterations for comparability, since our independent three-fold cross validation did not show signs of overfitting.

\subsection{Flow ablation experiments}
\label{ssec:flowAb}

Here we present an extensive number of ablations based on HD$^3$ to assess the quality of all our proposed contributions. We want to stress that all results in Tab.~\ref{tab:Ablation_FlyingChairs2} were obtained by \textbf{solely training on the \fc training set}. More specifically, we report error numbers (\epe and \pol; lower is better) and compare the original HD$^3$ model zoo baseline against our own, retrained baseline model, followed by adding combinations of our proposed contributions. We report performance on the target domains validation set (\fc), as well as on unseen data from different datasets (\ft, \sintel and \kitti), to gain insights on generalization behavior.
\begin{table*}[t]
	\centering
	\caption{Ablation results when training HD$^3$ CVr$\pm4$ on \fc v.s. the official model zoo baseline, our re-trained baseline and when adding all our proposed contributions. Results are shown on validation data for \fc and \ft (validation set used in the original HD$^3$ code repository), and on the official training data for \sintel, \kitT and \kitF, due to the lack of a designated validation split. (Highlighting \tbf{best} and \und{second-best} results).}
	\resizebox{\textwidth}{!}{
		\begin{tabular}{ccccc|cc|cccccccccc}
			\toprule
			\textsc{Gradient} & \textsc{Sampling} & \textsc{Flow} & \textsc{SAD} & \textsc{LMP} & \multicolumn{2}{c}{\fc}  &  \multicolumn{2}{c}{\ft}   & \multicolumn{2}{c}{\sintel final} & \multicolumn{2}{c}{\sintel clean} & \multicolumn{2}{c}{\kitT} & \multicolumn{2}{c}{\kitF} \\
			\textsc{Stopping} &                   & \textsc{Cues} &              &              &  \epe [1]   & \pol [\%]  &   \epe [1]   &  \pol [\%]  &  \epe [1]   &   \pol [\%]   &  \epe [1]   &   \pol [\%]   &  \epe [1]   &  \pol [\%]  &  \epe [1]   &  \pol [\%]  \\ \midrule
			\multicolumn{5}{l|}{HD$^3$ baseline model zoo}                                      &    1.439    &    7.17    &    20.973    &    33.21    &    5.850    &  \tbf{14.03}  &     3.70    & \tbf{8.56}    &   12.604    &    49.13    &    22.67    &    57.07    \\
			\multicolumn{5}{l|}{HD$^3$ baseline -- re-trained}                                  &    1.422    &    6.99    &    17.743    &    26.72    &    6.273    &  \und{15.24}  &     3.90    &     10.04     &    8.725    & \und{34.67} &    20.98    &    50.27    \\
			\cmark            &      \xmark       &    \xmark     &    \xmark    &    \xmark    &    1.215    &    6.23    &    19.094    &    26.84    &    5.774    &     15.89     &     3.72    &     10.51     &    9.469    &    44.58    &    19.07    &    53.65    \\
			\cmark            &      \xmark       &    \cmark     &    \xmark    &    \xmark    &    1.216    &    6.24    &    16.294    &    26.25    &    6.033    &     16.26     &     3.43    &      9.98     &    7.879    &    43.92    &    17.97    &    51.14    \\
			\cmark            &      \cmark       &    \xmark     &    \xmark    &    \xmark    &    1.208    &    6.19    &    17.161    &    24.75    &    6.074    &     15.61     &     3.70    &      9.96     &    8.673    &    45.29    &    17.42    &    51.23    \\
			\cmark            &      \cmark       &    \cmark     &    \xmark    &    \xmark    &    1.186    &    6.16    &    19.616    &    28.51    &    7.420    &     15.99     &     3.61    & \und{9.39}    & \tbf{6.672} & \tbf{32.59} & \tbf{16.23} &    47.56    \\
			\cmark            &      \cmark       &    \cmark     &    \cmark    &    \xmark    &    1.184    &    6.15    & \und{15.136} &    25.00    & \und{5.625} &     16.35     &\und{3.38}   &      9.97     &    8.144    &    41.59    &    17.13    &    52.51    \\ \hline
			\cmark            &      \xmark       &    \xmark     &    \xmark    &    \cmark    &    1.193    &    6.02    &    44.068    &    40.38    &   12.529    &     17.85     &     5.48    &     10.95     &    8.778    &    42.37    &    19.08    &    51.13    \\
			\cmark            &      \cmark       &    \cmark     &    \xmark    &    \cmark    & \und{1.170} & \und{5.98} &    15.752    & \und{24.26} &    5.943    &     16.27     &     3.55    &      9.91     &    7.742    &    35.78    &    18.75    & \tbf{49.67} \\
			\rowcolor{mapillarygreen}
			\cmark            &      \cmark       &    \cmark     &    \cmark    &    \cmark    & \tbf{1.168} & \tbf{5.97} & \tbf{14.458} & \tbf{23.01} & \tbf{5.560} &     15.88     &\tbf{3.26}   &      9.58     & \und{6.847} &    35.47    & \und{16.87} & \und{49.93} 
			\\ \bottomrule
	\end{tabular}}
	\label{tab:Ablation_FlyingChairs2}
	\vspace{-0.3cm}
\end{table*}

Our ablations show a clear trend towards improving \epe and \pol, especially on the target domain, as more of our proposed improvements are integrated. Due to the plethora of results provided in the table, we highlight some of them next. \textsc{\Detach} is often responsible for a large gap \wrt to both baseline HD$^3$ models, the original and our re-trained. Further, all variants with activated \textsc{Sampling} lead to best- or second-best results,  except for \pol on \sintel. Flow Cues give an additional benefit when combined with \textsc{Sampling} but not with warping. Another relevant insight is that our full model using all contributions at the bottom of the table always improves on \pol compared to the variant with deactivated \textsc{LMP}. This shows how \textsc{LMP} is suitable to effectively reduce the number of outliers by focusing the learning process on the under-performing (and thus more rare) cases. 

We provide additional ablation results on \ft and \sintel in  Tab.~\ref{tab:sintel_ablation}. 
The upper half shows \textsc{PreTrained} (P) results obtained after training on \fc and \ft, while the bottom shows results after additionally \textsc{fine-tuning} (F) on \sintel.
Again, there are consistent large improvements on the target domain currently trained on, i.e. (P) for \ft and (F) for \sintel. 
On the cross dataset validation there is more noise, especially for sintel final that comes with motion blur etc., but still always a large improvement over the baseline.
After finetuning (F) the full model with CVr$\pm8$ shows much better performance on sintel and at the same time comparable performance on \ft to the original baseline model directly trained on \ft.

\newcommand*\rot{\rotatebox{90}}
\begin{table*}[t]
  \centering
  \caption{Ablation results on \sintel, highlighting \tbf{best} and \und{second-best} results. Top: Baseline and \fc \& \ft pre-trained (P) models only. Bottom: Results after additional fine-tuning (F) on \sintel. }
\label{tab:sintel_ablation}
    \resizebox{0.99\textwidth}{!}{
    \begin{tabular}{ccccccc|cccccc}
\toprule
                        \textsc{Fine-tuned} & \textsc{Gradient} & \textsc{\textsc{Sampling}} & \textsc{Flow} & \textsc{SAD} & \textsc{LMP} & \textsc{CV} & \multicolumn{2}{c}{\ft}  & \multicolumn{2}{c}{\sintel final}  & \multicolumn{2}{c}{\sintel clean}  \\

\textsc{Pretrained} & \textsc{Stopping}& & \textsc{Cues} & & & \textsc{range $\pm8$} &\epe [1] & \pol&\epe [1] & \pol&\epe [1] & \pol\\
\midrule
\multicolumn{4}{c}{HD$^3$ baseline -- re-trained} & & & &12.52 &     18.06\% &     13.38 &        16.23 \% &         3.06 &    6.39\% \\
\multicolumn{1}{c}{P} &        \cmark &     &  &  &      &     &  7.98 &     13.41\% &      \tbf{4.06} &        \tbf{10.62} \% &         \und{1.86} &   \und{ 5.11}\% \\
\multicolumn{1}{c}{P} & \cmark & \cmark & \cmark & \cmark & \cmark & &  \und{7.06} & \und{12.29}\% & \und{4.23} & \und{11.05} \% & 2.20 &    5.41\% \\
\multicolumn{1}{c}{P} & \cmark &   \cmark &  \cmark &  \cmark & \cmark & \cmark &  \tbf{5.77} & \tbf{11.48}\% & 4.68 & 11.40 \% &  \tbf{1.77} & \tbf{4.88}\% \\
 \midrule
\multicolumn{1}{c}{F} & &     &  &  &      &     & 19.89 &       27.03\% &      (1.07) & (4.61 \%) &  1.58 &    4.67\% \\
F & \cmark &     &  &  &      &     &    \und{13.80} &       \und{ 20.87}\% &      (0.84) &  (3.79 \%) &  \und{1.43} &    4.19\% \\
F & \cmark & \cmark &  \cmark &  \cmark & \cmark & & 14.19 & 20.98\% & \und{(0.82)} &  \und{(3.63 \%)} &  \und{1.43} &    \und{4.08}\% \\
       \rowcolor{mapillarygreen}
F & \cmark & \cmark &  \cmark &  \cmark & \cmark & \cmark & \tbf{11.80} & \tbf{19.12}\% & \tbf{(0.79)} &  \tbf{(3.49 \%)} &  \tbf{1.19} &    \tbf{3.86}\% \\
\bottomrule
\end{tabular}
}
% \end{adjustbox}
\end{table*}

\subsection{Optical flow benchmark results}
\label{ssec:benchmarks}
The following provides results on the official \sintel and \kitti test set servers.

\paragraph{\sintel.} 
By combining all our contributions and by using a cost volume search range of $\pm8$, we set a new state-of-the-art on the challenging \sintel \textsc{Final} test set, improving over the very recent, best-working approach in~\cite{BarHaim_2020_CVPR} (see Tab.~\ref{tab:sintel_test}). Even by choosing the default search range of CVr$\pm4$ as in~\cite{HD3Flow_yin2019hd3} we still obtain significant improvements over the HD$^3$-ft baseline on training and test errors.

\paragraph{\kitT and \kitF.} We also evaluated the impact of our full model on \kitti and report test data results in Tab.~\ref{tab:kitti_aep_error}. We obtain new state-of-the-art test results for \epe and \pol on \kitT, and rank second-best at \pol on  \kitF. 
On both, \kitT and \kitF we obtain strong improvements on the training set on \epe and \pol.
Finally, while on \kitF the recently published VCN~\cite{VCN_NIPS2019_8367} has slightly better \pol scores, we perform better on foreground objects (test Fl-fg 8.09~\%  vs. 8.66~\%) and generally improve over the HD$^3$ baseline (Fl-fg 9.02~\%).
It is worth noting that all \kitti finetuning results are obtained after integrating knowledge distillation from \S~\ref{ssec:AdditionalRefinements}, leading to significantly improved flow predictions on areas where \kitti lacks training data (\eg in far away areas including sky, see Fig.~\ref{fig:vis_comparison_distilled_nondistilled}). We provide further qualitative insights and direct comparisons in the supplementary material (\S~C).

\begin{table}[t]
  \centering
	\begin{minipage}[t]{.42\linewidth}
	
    \caption{ \epe scores on the \sintel test datasets. The appendix -\textit{ft} denotes fine-tuning on \sintel. }
    \label{tab:sintel_test}
    \resizebox{0.958\columnwidth}{!}{
    \begin{tabular}{c|cc|cc}
    	\toprule
    	                                  & \multicolumn{2}{c}{\textsc{Training}} & \multicolumn{2}{c}{\textsc{Test}} \\ \midrule
	           \textsc{Method}            & \textsc{Clean} &    \textsc{Final}    & \textsc{Clean} &  \textsc{Final}  \\ \midrule
	      FlowNet2~\cite{IMSKDB17}        &      2.02      &         3.14         &      3.96      &       6.02       \\
	     FlowNet2-ft~\cite{IMSKDB17}      &     (1.45)     &        (2.01)        &      4.16      &       5.74       \\
	    PWC-Net~\cite{PWCNetSun2018}      &      2.55      &         3.93         &       -        &        -         \\
	   PWC-Net-ft~\cite{PWCNetSun2018}    &     (2.02)     &        (2.08)        &      4.39      &       5.04       \\
	   SelFlow~\cite{SelFlow_Liu2019}     &      2.88      &         3.87         &      6.56      &       6.57       \\
 	  SelFlow-ft~\cite{SelFlow_Liu2019}   &     (1.68)     &        (1.77)        &      3.74      &       4.26       \\
   	IRR-PWC-ft~\cite{IRR_PWC_Hur2019CVPR} &     (1.92)     &        (2.51)        &      3.84      &       4.58       \\
  	 PWC-MFF-ft~\cite{MFF_ren2018fusion}  &       -        &          -           & \und{3.42}     &       4.56       \\
	   VCN-ft~\cite{VCN_NIPS2019_8367}    &     (1.66)     &        (2.24)        & \tbf {2.81}    &       4.40       \\
  	   ScopeFlow \cite{BarHaim_2020_CVPR} &       -        &         -            &      3.59      &  \und{4.10}       \\ 
  	   Devon \cite{Lu_2020_WACV_Devon}    &      -         &         -            &      4.34      &      6.35         \\ \midrule 
  	 HD$^3$~\cite{HD3Flow_yin2019hd3}     &      3.84      &         8.77         &       -        &        -         \\
  	 HD$^3$-ft~\cite{HD3Flow_yin2019hd3}  &     (1.70)     &         (1.17)      &      4.79      &       4.67       \\ \midrule
  	   	\rowcolor{mapillarygreen}
    IOFPL-no-ft                            &      2.20      &   4.32               &      -     &  -  \\
   	\rowcolor{mapillarygreen}
                                  IOFPL-ft & \und{1.43}     &   \und{(0.82)}       &      4.39      &   4.22   \\ \bottomrule
   	\rowcolor{mapillarygreen}
                                  
                        IOFPL-CVr8-no-ft    &      1.77      &   4.68               &      -         &  -  \\
   	\rowcolor{mapillarygreen}
                              IOFPL-CVr8-ft & \tbf{1.19}     &   \tbf{(0.79)}       &      3.58     &  \textbf{4.01}   \\ \bottomrule
    \end{tabular}
	}
	\end{minipage}
	~~\begin{minipage}[t]{.55\linewidth}
    \caption{\epe and \pol scores on the \kitti test datasets. The appendix -\textit{ft} denotes fine-tuning on \kitti. Ours is IOFPL.} 
    \label{tab:kitti_aep_error}
    \resizebox{0.99\columnwidth}{!}{
    \begin{tabular}{c|ccc|ccc}
    \toprule
      & \multicolumn{3}{c}{\textsc{\kitT}}& \multicolumn{3}{c}{\textsc{\kitF}} \\
    \midrule
    \textsc{Method}                      & \epe & \epe & \pnoc [\%]& \epe  & \pol  [\%]& \pol [\%]\\
                                         & train   & test    & test   & train  & train & test   \\
    \midrule
    FlowNet2~\cite{IMSKDB17}             &  4.09   & -      &  -     & 10.06   & 30.37  & - \\
    FlowNet2-ft~\cite{IMSKDB17}          &  (1.28) & 1.8    &  4.82  & (2.30)  & 8.61   & 10.41\\
    PWC-Net~\cite{PWCNetSun2018}         &  4.14   & -      &  -     & 10.35   & 33.67  & - \\
    PWC-Net-ft~\cite{PWCNetSun2018}      &  (1.45) & 1.7    &  4.22  & (2.16)  & 9.80   & 9.60\\
    SelFlow~\cite{SelFlow_Liu2019}       &  1.16   & 2.2    &  7.68  & (4.48)  &  -     & 14.19\\
    SelFlow-ft~\cite{SelFlow_Liu2019}    &  (0.76) & 1.5    &  6.19  & (1.18)    & -      & 8.42\\
    IRR-PWC-ft~\cite{IRR_PWC_Hur2019CVPR}&  -      & -      &  -     & (1.63)    &  5.32  & 7.65\\
    PWC-MFF-ft~\cite{MFF_ren2018fusion}  &  -      & -      &  -     & -       &  -     & 7.17\\
    ScopeFlow \cite{BarHaim_2020_CVPR}   &  -      & 1.3    &  2.68  & -       & -      & 6.82    \\
    Devon \cite{Lu_2020_WACV_Devon}      &  -      & -      &  6.99  & -       &        & 14.31	  \\
    VCN~\cite{VCN_NIPS2019_8367}         &  -      & -      &  -     & \und{(1.16)}     & 4.10    & \tbf{6.30}\\    
    \midrule
    HD3F~\cite{HD3Flow_yin2019hd3}       &  4.65   & -      &  -     & 13.17   & 23.99  & \\
    HD3F-ft~\cite{HD3Flow_yin2019hd3}    &  (0.81) & \und{1.4} &   \tbf{2.26}  & 1.31   &  4.10 & 6.55\\
    \midrule
    \rowcolor{mapillarygreen}
    IOFPL-no-ft                           &   2.52  & -            & -           & 8.32        & 20.33         & -
    \\
  	\rowcolor{mapillarygreen}
    IOFPL-ft                              &  \tbf{(0.73)}      & \tbf{1.2} & \und{2.29}   & 1.17  & 3.40  & 6.52 \\
    \rowcolor{mapillarygreen}
    IOFPL-CVr8-no-ft                        &    2.37  & -      & -      & 7.09        & 18.93         & -
    \\
  	\rowcolor{mapillarygreen}
    IOFPL-CVr8-ft                           &  \und{(0.76)}      & \tbf{1.2} & \tbf{2.25}  & \tbf{1.14}  & \tbf{3.28}  & \und{6.35} \\
    \bottomrule
    \end{tabular}	
	}
	\end{minipage}
\end{table}

%------------------------------------------------------------------------

\section{Conclusions}
In this paper we have reviewed the concept of spatial feature pyramids in context of modern, deep learning based optical flow algorithms. We presented complementary improvements for cost volume construction at a single pyramid level, that i) departed from a warping- to a sampling-based strategy to overcome issues like handling large motions for small objects, and ii) adaptively shifted the focus of the optimization towards under-performing predictions by means of a loss max-pooling strategy. We further analyzed the gradient flow across pyramid levels and found that properly eliminating noisy or potentially contradicting ones improved convergence and led to better performance. We applied our proposed modifications in combination with additional, interpretable flow cue extensions as well as distillation strategies to preserve knowledge from (synthetic) pre-training stages throughout multiple rounds of fine-tuning. We experimentally analyzed and ablated all our proposed contributions on a wide range of standard benchmark datasets, and obtained new state-of-the-art results on \sintel and \kitT.

\paragraph*{Acknowledgements:}
T. Pock and M. Hofinger acknowledge that this work was supported by the ERC starting grant HOMOVIS (No. 640156).

%% file: supmat_content.tex
This document contains supplementary material for the paper 'Improving Optical Flow on a Pyramid Level'.
The structure of this supplementary document is the following:
\begin{itemize}
	\item Further insights and experiments on gradient stopping:
	\begin{itemize}
		\item Variance analysis
		\item Smoothness analysis
		\item Cross-task evaluation - stereo
		\item Cross-architecture evaluation - PWC
	\end{itemize}

	\item Details on Flow Cues:
	\begin{itemize}
		\item Notation
		\item Detailed explanation
	    \item Ablations on flow cues
	\end{itemize}

	\item Further Analysis: 
	\begin{itemize}
		\item Ablation on Data distillation 
		\item Ablation on extending search range
		\item Qualitative comparisons of validation epe changes
		\item Histograms of errors
		\item Qualitative training results for KITTI (images)
		\item Qualitative training results for MPI-Sintel (images)
		\item Qualitative Results on KITTI
		\item Qualitative Results on MPI-Sintel
		\item Sidenote on D2V and V2D operation with warping vs. sampling
	\end{itemize}
\end{itemize}

\section{Further insights on Gradient Stopping}
\label{sec:grad_intro}
In this section we provide additional empirical insights to what is described in Section 3.3 in the main submission document, \ie, why stopping the optical flow gradients between the levels is beneficial. We will do this by showing that the variance of the model parameter gradients over the training set is reduced (sec~\ref{sec:grad_var}), and the Lipschitzness of the parameter gradients is improved as well, while still providing a descent direction (sec~\ref{sec:grad_lipschitz}). For stochastic gradient descent methods these properties lead to improved convergence, which is what we already observed in the main paper (Fig. 5).
Finally, we show that gradient stopping also leads to improved convergence for the different task of stereo estimation (sec~\ref{sec:grad_stereo}) as well as for a different architecture like PWC-Net (implemented in a different code-base (sec~\ref{sec:grad_PWC} ).

\subsection{Gradient stopping - variance analysis}
\label{sec:grad_var}
\begin{figure}[h]
    \centering
    \includegraphics[width=\columnwidth]{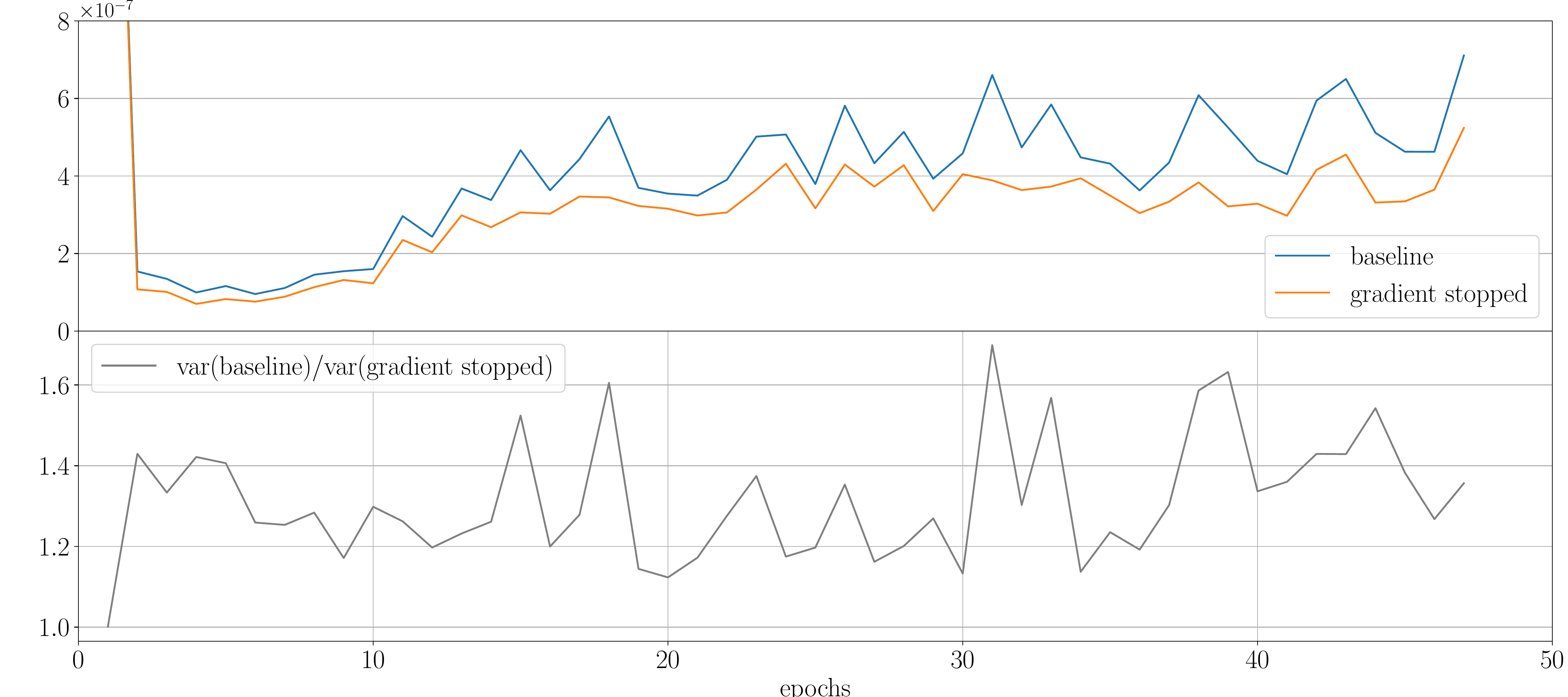}
    \caption{Gradient variance for a HD\textsuperscript{3} baseline model vs. a model with the proposed gradient stopping. The baseline has a higher gradient variance over the training data, which leads to slower convergence.}
   \label{fig:grad_var}
\end{figure}

It is known \cite{Wang_NIPS2013_5034_VarianceSGD} that for stochastic gradient descent methods the rate of convergence decreases with increasing variance of the gradients over the training set. We can show empirically that stopping the optical flow gradients between levels (see Fig.~4 in the main paper) leads to a reduced variance of the gradients w.r.t~the whole training dataset when compared to the baseline model.
To ensure a fair and valid comparison, both model versions use identical parameters $\Theta$ and are fed with the exact same data batches $\xi_n$ all the time. 
The variance over each epoch is computed independently for every single parameter using  Welford's online variance computation algorithm~\cite{Welford1962_OnlineVariance} in a numerically stable variant. 
After each epoch, the mean of these $P$ single parameter variances is computed for each model as
\begin{align}
\sigma ^2 = \frac{1}{P}  \sum_{\theta \in \Theta}   \underset{n \in N}{ \text{VAR}_\text{Welford}}(\nabla f_\theta( \xi_n))
\end{align}
and shown in Fig.~\ref{fig:grad_var}.
After gradient variances are computed a standard training is performed for 1 epoch, and the parameters of both models are update with the new parameters of the baseline model to ensure that the only difference in the gradient variance comes from the gradient computation itself.
As can be seen in Fig.~\ref{fig:grad_var} our proposed partial gradient stopping truly reduces the gradient variance w.r.t the model parameters and the training dataset.
This leads to the improved rate of convergences for our proposed partial gradient stopping over the baseline.

\subsection{Gradient stopping - smoothness analysis}
\label{sec:grad_lipschitz}
%"'Effective' refers here to measuring the change of gradients as we move in the direction of the gradients"
%https://papers.nips.cc/paper/7515-how-does-batch-normalization-help-optimization.pdf

\begin{figure}[hbtp]
	\centering
	\includegraphics[width=\columnwidth]{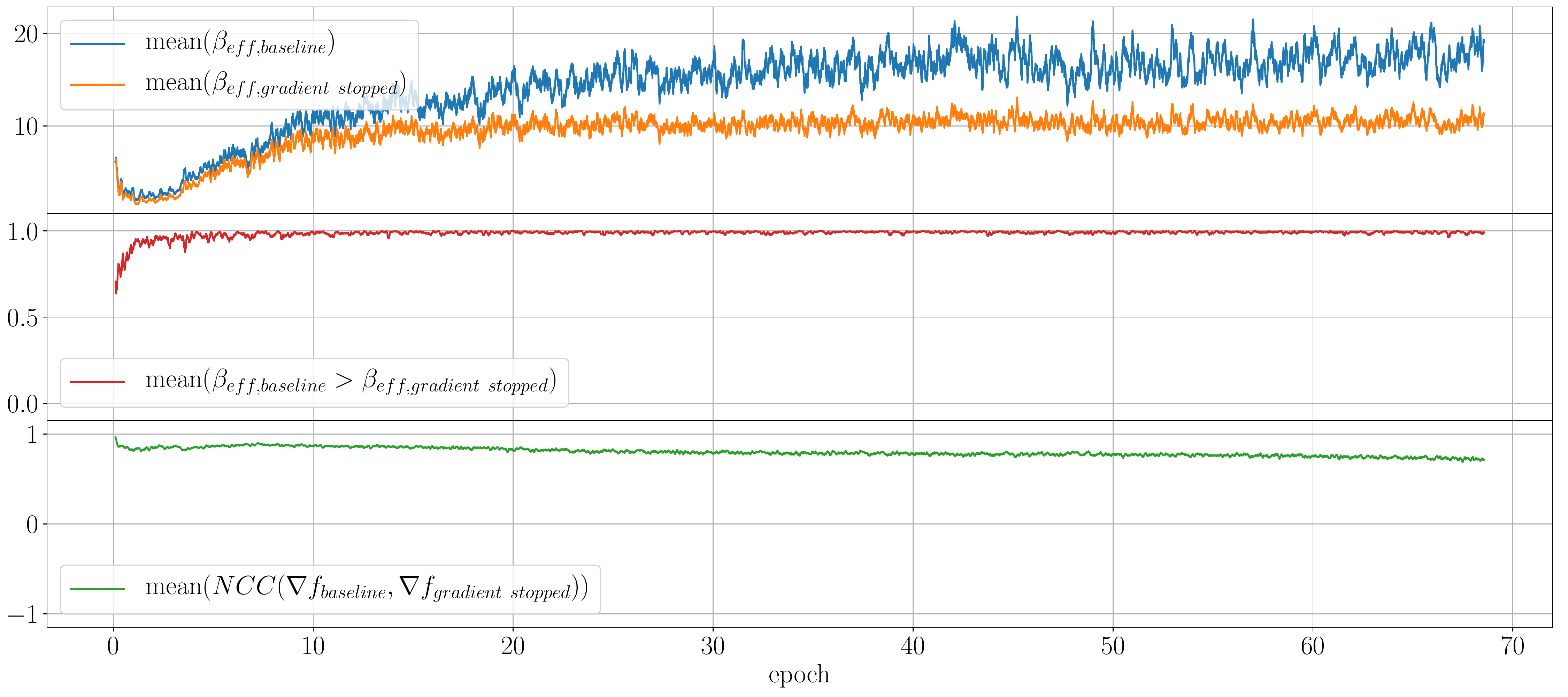}
	\caption{Partial gradient stopping vs. Lipschitzness of gradients. Top: Average of the effective $\beta$-smoothness shows that model with gradient stopping is smoother (lower $\beta_{eff}$) than the baseline; Middle: Percentage of how often gradient stopping leads to smoother results; Bottom: Positive normalized cross correlation between the model parameter gradients indicates that it is still a descent direction.  }
	\vspace{-1em}
	\label{fig:grad_lipschitz}
\end{figure}

Here we will show that stopping the partial optical-flow gradient between the levels also leads to a better Lipschitzness of the gradients of the loss also known as $\beta$-smoothness, while still providing a descent direction. It is well known that the rate of convergence increases if the function has a low curvature which corresponds to a low  $\beta$-smoothness.  We follow the approach of \cite{Nips2018_Santurkar_Batchnorm} that estimate 'effective' $\beta$-smoothness ($\beta_\text{eff}$) by measuring the $l_2$ gradient change over difference in parameters, as they move along the gradient direction in the optimization. 
\begin{align}
\label{eq:lipschitz_const}
\beta_{\text{eff}} &= \frac{\| \nabla f(\xi,\Theta_1) - \nabla f(\xi,\Theta_2) \|_2 }{\| \Theta_1 - \Theta_2 \|_2 }
\end{align}
We ensure a fair comparison between the baseline model and our version with partial flow gradient stopping by evaluating the gradient functions with the exact same parameters $\Theta_i$ and data batches $\xi_n$ for both model versions at all times.
Fig.~\ref{fig:grad_lipschitz} 
(top) shows that on average  $\beta_{\text{eff}}$ is lower which corresponds to a lower curvature. This is confirmed by the center plot that directly compares $\beta_{\text{eff}}$ for both models in every iteration before averaging the result. The lower plot shows that the normalized cross correlation (NCC) of the gradients for the parameters of both models are positively correlated. This is in contrast to the NCC of the partial optical flow gradients (Fig. 5, main paper) between the levels.
Therefore, stopping the partial optical flow gradients between the levels, reduces intermediate parts that oppose each other, which in turn leads to better final gradients at the model parameters. The latter are still positively correlated with the original parameter gradients, which shows that they still provide a descent direction, but with better convergence properties, as shown by our various insights. Finally, based on these analyses and the improved results obtained in our experimental section we conclude the importance of blocking partial optical flow gradients across levels in a pyramidal setting for improved convergence.

\subsection{ Gradient stopping on depth estimation -- HD$^3$, depth from Stereo.}
\label{sec:grad_stereo}
With this experiment we show that stopping the partial optical flow gradients between the levels also works for stereo estimation.
We use the Stereo training setup of \HDq~in their original publicly available codebase\footnote{HD$^3$ codebase : \url{https://github.com/ucbdrive/hd3/}} and run a training on the \ft Stereo dataset. We choose to use the original version of the code base just with \detach added, and keep the original training procedure that trains only on the  \textit{left} disparity. We do this to show that the effect of \detach is not just limited to the simultaneous forward and backward training used in the main paper, but is a more general one.

\begin{figure}[tbph]
    \centering
    \includegraphics[width=\columnwidth]{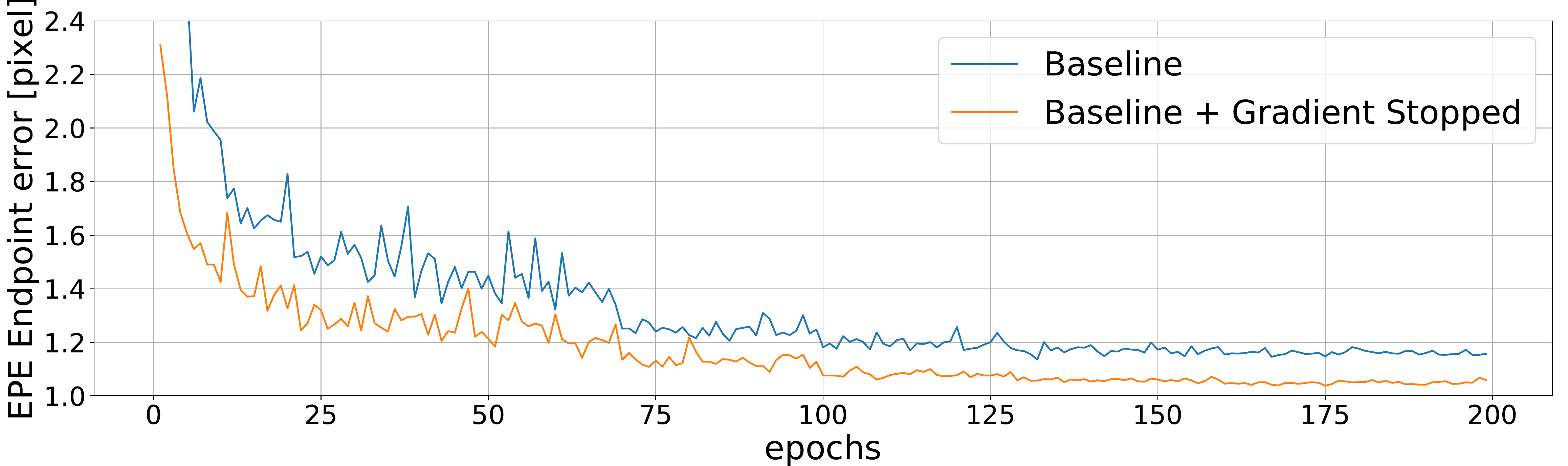}
    \caption{Improving HD$^3$ Stereo estimation with \detach. Curves show validation Endpoint error (EPE) after each training epoch. Simple gradient stopping leads to faster convergence of the \epe}
    \vspace{-1em}
   \label{fig:LossCurve_Stereo}
\end{figure}

Again, we find significant improvements with our proposed partial flow gradient stopping, as can be seen in Fig. ~\ref{fig:LossCurve_Stereo}, which leads to an improvement of $\approx$10\% on the final \epe.
This confirms that \detach also works for stereo estimation networks.
Furthermore, it verifies that \detach does not require joint forward- and backward flow training as used in the flow ablations in the main paper, but also leads to significant gains for a standard forward-only training.

\subsection{ Gradient stopping on different architectures estimation -- Improving PWC-Net Optical Flow.}
\label{sec:grad_PWC}
With this experiment we show that this behaviour is not limited to HD\textsuperscript{3} but also applies to other networks as PWC.
We use the PWC-Net implementation from the official IRR-PWC~\cite{IRR_PWC_Hur2019CVPR} publicly available code base\footnote{IRR-PWC codebase: \url{https://github.com/visinf/irr}}, and run a training on the \fcOne dataset using their provided data processing and augmentation strategy, and follow all default settings for training. We run two experiments, the baseline and an experiment where we apply \detach at the upsampling layer within the pyramid structure used therein.
In direct comparison we found both, significantly improved reduction of the training loss for the final high-resolution level as well as the validation \epe (\pol is not reported from their inference code). 

\begin{figure}
    \centering
    \includegraphics[width=\columnwidth]{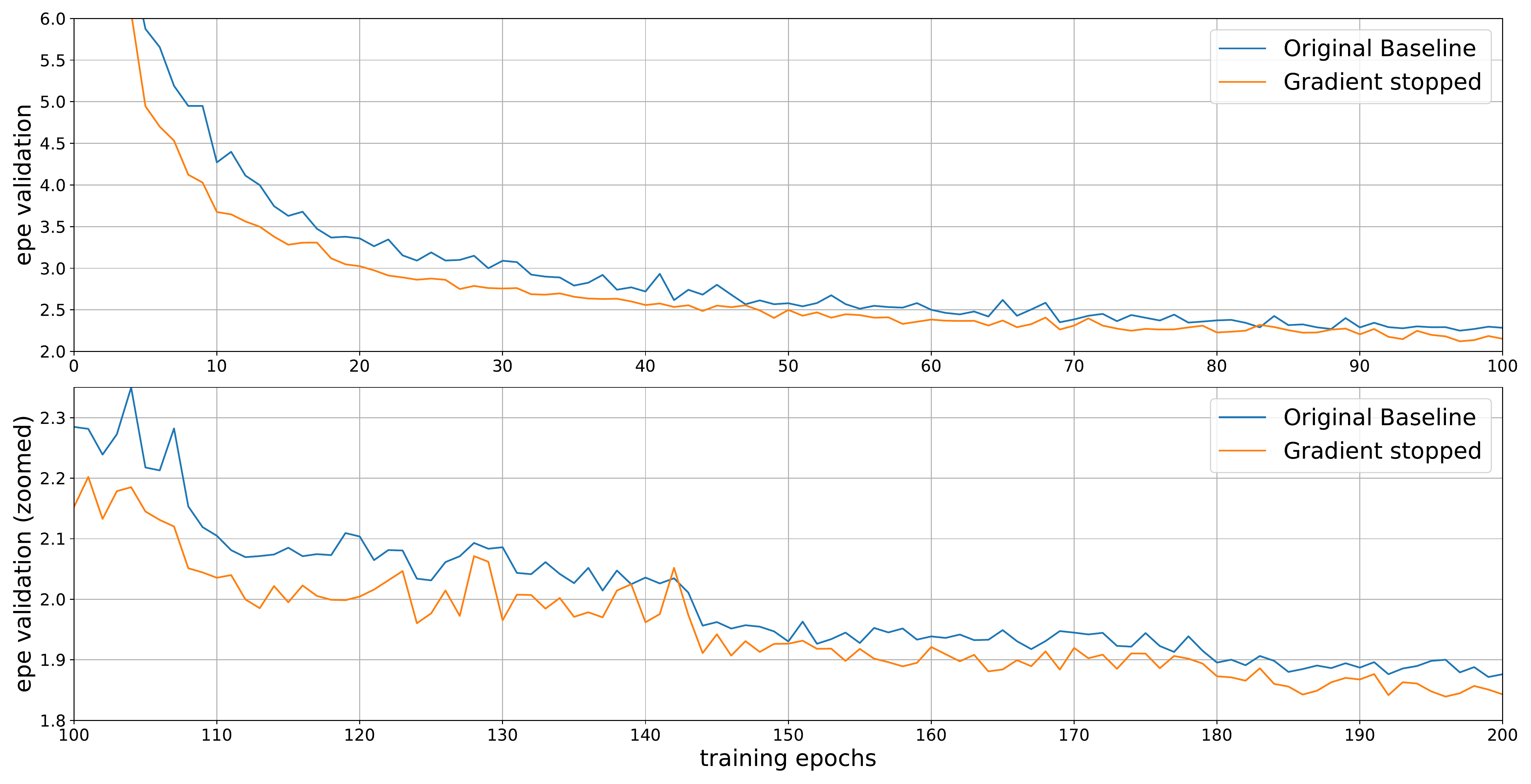}
    \caption{Improving PWC-Net with \detach. Training with gradient stopping vs. original. Gradient stopping leads to faster decrease for the validation EPE. 
    }
    \vspace{-1em}
    \label{fig:PWC_Epe}
\end{figure}

Fig.~\ref{fig:PWC_Epe} shows the validation \epe of an exemplary experimental result on the PWC flow Network. As can be seen, applying \detach  leads to a faster convergence of the \epe.
This immediately leads to initial gains of  more than 10\% at 20 epochs and 6\% at 100 epochs.
Therefore, lower \epe values can be reached faster.
We kept the original learning rate schedule for comparability,  but even in this setting that was optimized for the original baseline, a difference of approximately 2\% remains after 200epoch.
\Detach shows a clear positive impact, even though the used PWC-variant directly regresses the flow at each level, whereas the HD$^3$ baseline that was used for many comparisons in the main paper uses residual estimates together with the D2V and V2D operations.
This shows that stopping the gradients for the flow at the upsampling layer leads to a faster decrease of the \epe also across multiple types of optical flow networks.

\section{Further details on Flow Cues}
\subsection{Notation}
To simplify equations in the following section, we define a few additional terms on top of the main paper.
Given a pixel $x\in \set I^l_1$ we denote by $x_{1\to 2}\in\mathbb R^2$ the matching position of $x$ in $I^l_2$ (in absolute terms), \ie $x_{1\to 2}=x+F^l_{1\to 2}(x)$.
Similarly, for the opposite direction, we define $y_{2\to 1}\in \mathbb R^2$ for pixels $y\in \set I_2$.

\subsubsection{Details on the Flow Cues module}
The use of prior knowledge when computing optical flow has been widely explored in classical methods.
Recently, \cite{IRR_PWC_Hur2019CVPR} successfully used forward-backward flow warping as feature for occlusion upsampling. 
Although this feature is hand crafted it is very valuable, as it provides cues that would otherwise be hard to learn for a convolutional network since it can connect completely different locations on the coordinate system of $I_1$ and $I_2$.
Classic approaches like inverse flow estimation  \cite{InverseFlow_Javier2013_Techreport} show that there are even more cues that can potentially be of interest.
We therefore propose to combine multiple of these cues, which can be mutually beneficial, and make them explicitly available to the network as cheaply computable features to directly improve flow predictions.

In order to do so, our architecture keeps jointly track of the forward and backward flows by exploiting Siamese modules with shared parameters, with features from $I_1$ and $I_2$ being fed to the two branches in mirrored order. 
A downside is its increased memory consumption, which we noticeably mitigate by adopting In-Place Activated BatchNorm~\cite{RotPorKon18a} throughout our networks.
Without additional connections, the Siamese modules compute the forward $F_{1\to 2}$ and backward $F_{2\to 1}$ flow mappings in a completely independent way.
However, in practice the true flows are strongly tied to each other, although they reside on different coordinate systems.
We therefore provide the network with a Flow Cue Module that gives each branch different kind of cues about its own and the other branch's flow estimates.
Each of these cues represents a different mechanism to bring mutually supplementary information from one coordinate system to the other.
For the sake of simplicity, we will always present the results of the cues in the coordinate system of the branch that operates on the features of $I_1$.

%%%%%%%%%%%%%%%%%%%%%%%%
% FwdBwd
\paragraph{Forward-backward flow warping.}
Since both flow mappings are available, they can be used to bring one flow in the coordinate frame of the other via dense warping.
For example, a forward flow estimate $F^{\text{fb}}_{1\to 2}$ can be made from the backward flow $F_{2\to 1}$ by warping it with the forward flow  $F_{1\to 2}$:
\begin{equation}
F_{1\to 2}^\text{fb}(x)=-F_{2\to 1}(x_{1\to 2})
%F_{2\to 1}^\text{fb}(y)=-F_{1\to 2}(y_{2\to 1})
\end{equation}
The other direction $F_{2\to 1}^\text{fb}$ can be computed in a similar way.
Comparing the estimated results $F_{1\to 2}^\text{fb}$ versus $F_{1\to 2}$ can be used for consistency checks, and is used in unsupervised flow methods ~\cite{SelFlow_Liu2019,UnsupFlow2016jjyu} to estimate occlusions in a heuristic manner.

%%%%%%%%%%%%%%%%%%%%%%%%
% Reverse Flow
\paragraph{Reverse flow estimation~\cite{InverseFlow_Javier2013_Techreport}.}
In contrast to the previous cue, reverse flow estimation can be used to estimate the forward flow $F_{1\to 2}$  directly from backward flow  $F_{2\to 1}$ alone, although in a non-dense manner. The reverse flow estimates are denoted by $F_{2\to 1}^\text{rev}$ and $F_{1\to 2}^\text{rev}$ and are obtained by 
\begin{equation}\label{eq:ffflow}
F^\text{rev}_{1\to 2}(x)=-\frac{\sum_{y\in \set I_2} \omega(x,y_{2\to 1}) F_{2\to 1}(y)}{\omega_1(x)}\,,
\end{equation}
where $\omega(x,x')=[1-|x_u-x'_u|]_+  \  [1-|x_v-x'_v|]_+$ denotes the bilinear interpolation weight of $x'$ relative to $x$, and
\begin{equation}
\omega_1(x)=\sum_{y\in \set I_2}\omega(x,y_{2\to 1})\,
\end{equation}
is a normalizing factor.
In the dis-occluded areas where the denominator of Eq.~\eqref{eq:ffflow} is 0, we define the flow values  $F^\text{rev}_{1\to 2}(x)=0$.
In occluded areas  $F^\text{rev}_{1\to 2}$ will become an average of the incoming flows.
Similarly, we define $F^\text{rev}_{2\to 1}$ by swapping $1$ and $2$ as well as $x$ and $y$ in Eq.~\eqref{eq:ffflow}.

%%%%%%%%%%%%%%%%%%%%%%%%
% Map uniqueness density
\paragraph{Map uniqueness density~\cite{Unger2012,OccAwareUnsupLearning__Wang2018_CVPR}.}  Provides information about occlusions and dis-occlusions and basically corresponds to $\omega_1$ in Eq.~\eqref{eq:ffflow} for image $I_1$.
The value of $\omega_1(x)$ provides the (soft) amount of pixels in $I_2$ with flow vectors pointing towards $x\in \set I_1$.
Occluded areas will result in values $\geq 1$ whereas areas becoming dis-occluded in values $\leq 1$. $\omega_1$  is therefore an indicator on where the reverse flow is more or less precise.
Similarly, we have $\omega_2$(x) for $I_2$.

%%%%%%%%%%%%%%%%%%%%%%%%
% Out of Image occlusions
\paragraph{Out-of-image occlusions.} This represents an indicator function, \eg $o_1: \set I_1\to\{0,1\}$ for image $I_1$, providing information about flow vectors pointing out of the other image's domain, \ie
\begin{equation}
%o_2(y)=\mathbb 1_{y_{2\to 1}\notin \set I_1}
o_1(x)=\mathbb 1_{x_{1\to 2}\notin \set I_2}
\end{equation}
and similarly we define $o_2: \set I_2 \to \{0,1\}$ for image $I_2$.

\begin{figure}
	\centering
	\includegraphics[width=\columnwidth]{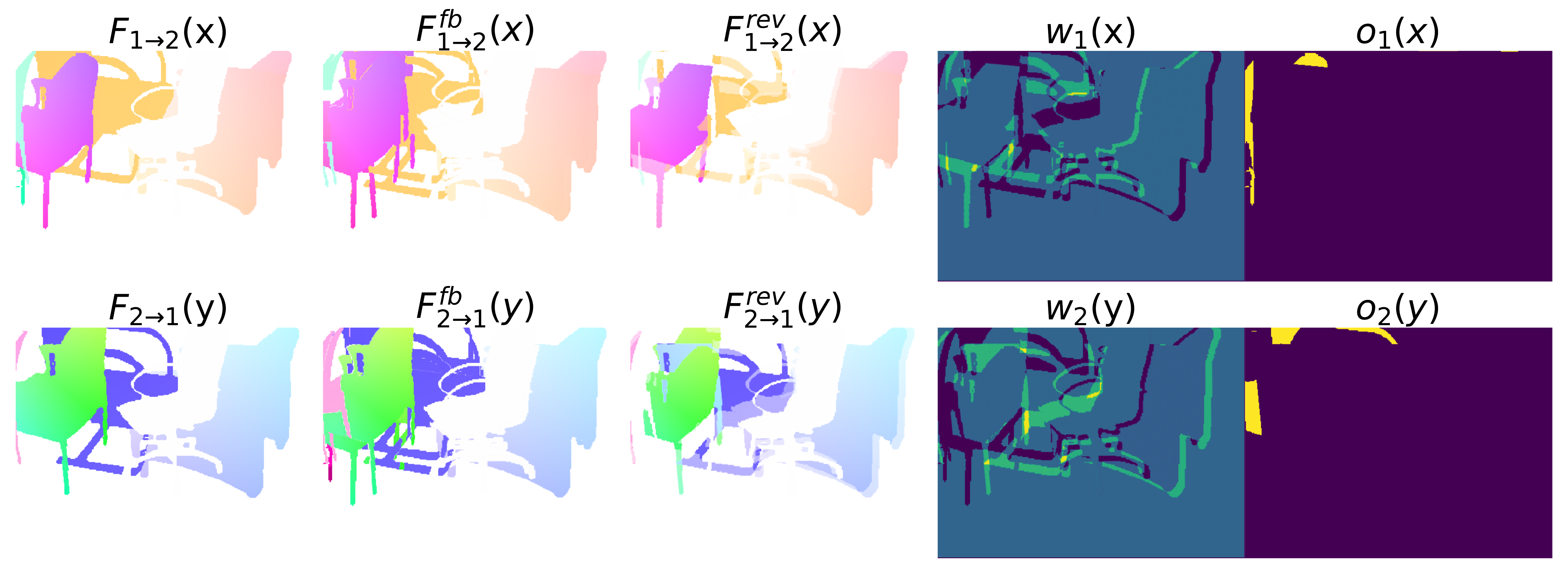}
	\vspace{-1em}
	\caption{Flow Cues module output illustration for a given optical flow input; Left to right: Input flow, forward-backward estimate, reverse flow estimate, map uniqueness density, out of image occlusions. Note the differences in the $F^{\text{fb}}$ and $F^{\text{rev}}$ and how }
	\vspace{-1em}
	\label{fig:FlowCues}
\end{figure}

\paragraph{The Flow Cue Module.}
We show in Fig.~\ref{fig:FlowCues} how the flow cues mutually benefit from one another in different areas. 
\Eg, the-out-of-image occlusions $o_1$ allow to differentiate which dis-occlusions in map uniqueness density $\omega_1$ are real dis-occlusions, \ie areas where the object moved away, and where the low density stems from flow vectors in the second image that are just likely not visible in the current crop.

We therefore provide the network with all the additional flow cues mentioned above, by stacking them as additional features together with the original forward flow $F_{1\to 2}$ for the subsequent part of the network.
Therefore, the network now has three differently generated flow estimates including its own prediction $F_{1\to2}$. The following layers can therefore reason about consistency and probable sources of outliers with a far better basis than one single cue alone could provide.
Symmetrically, the same is done for the backward stream (see Fig.~\ref{fig:FlowCues}).

\subsection{Ablations on Contributions of Flow Cues.}
\label{ssec:FlowCues}

Here we evaluate the impact of our proposed flow cues in comparison to related ones from prior works~\cite{IRR_PWC_Hur2019CVPR,FlowNet3_Ilg2018}, demonstrating their effect on relevant error measures on the \fc dataset.
The ablations are performed training on Flying Chairs 2. We use averages over the last 10 validation results to reduce the effect of single spikes.
In Tab.~\ref{tab:AblationsFlowCues} we list our findings, always on top of activating  \detach and \textsc{Sampling} due to its preferable behavior for estimating flow of fine-grained structures.

Providing \textit{Mapping Occurrence Density} (\textsc{MOD}) \cite{Unger2012,OccAwareUnsupLearning__Wang2018_CVPR}  as the only Flow Cue and hence information about the occlusions and dis-occlusions slightly degrades results in terms of both, \epe and \pol. When running the Sampling in combination with Forward-Backward flow warping (\textsc{FwdBwdFW}) we encounter a considerable reduction of errors -- particularly on the \pol errors. Finally, when combining \textsc{Sampling} with all our proposed Flow Cues  (\textsc{All Cues}), \ie reverse flow estimation, mapping occurrence density, and out-of-image occlusions, we obtain the lowest errors. 

\begin{table}
	\centering
	\caption{Ablation results on Flow Cues on top of Cost Volume Sampling and Gradient Stopping using CV-range of $\pm$4 pixels}
	\label{tab:AblationsFlowCues}
	\resizebox{0.5\textwidth}{!}{
		\begin{tabular}{ccc|cc}
			\toprule
			\rotatebox{0}{\textsc{MOD}} & \rotatebox{0}{\textsc{FwdBwdFW}} & \rotatebox{0}{\textsc{All Cues}} & \epe & \pol \\
			\midrule
			\xmark & \xmark & \xmark & 1.208 & 6.192 \\
			\cmark & \xmark & \xmark & 1.217 & 6.271 \\
			\xmark & \cmark & \xmark & 1.202 & 6.171 \\
			\rowcolor{mapillarygreen}
			\cmark & \cmark & \cmark & 1.186 & 6.156 \\
			\bottomrule
	\end{tabular}}
\end{table}

\section{Further Analysis}
\subsection{Details on Distillation}
\label{sec:distillation_details}

In this section we provide additional details on our distillation strategy.
In contrast to \cite{Hin+15} we don't want to transfer knowledge from a larger network or ensembles to smaller ones, but to transfer it from one domain to the other.
We therefore avoid to keep all predictions, since some are completely off, and instead try to filter out the most trustworthy.
Specifically, we apply the following filters, obtaining ``pseudo ground-truth'' annotations
(Fig.~6 main paper, bottom right):
\begin{itemize}
		\item We use forward $F_{1\to 2}$ and backward $F_{2\to 1}$ flows to estimate occlusions. Specifically, we regard a pixel $y\in I_1$ as not occluded if the following holds~\cite{SelFlow_Liu2019}
		\begin{multline}
		\hspace{-0.4cm}
		\Vert F_{1\to 2}(y)+F_{2\to 1}(y_{1\to 2})\Vert^2-0.05<0.01\left(\Vert F_{1\to 2}(y)\Vert^2 + \Vert F_{2\to 1}(y_{1\to 2})\Vert ^2  \right)
		\hspace{-0.1cm}
		\end{multline}
		\item We compute the photometric error using SAD on a per pixel basis and determine a mask of good predictions by thresholding the error.
		\item We determine the confidence of the network using the method proposed in~\cite{HD3Flow_yin2019hd3} and retain predictions with a confidence above $95\%$.
		\item We filter pixels that are more than 3 pixels away from the gt %\markus{This is used for 1 of 2 experiments} 
		\item Finally, we combine all of the previous filters and apply an additional pruning using an erosion operation to remove small patches, in order to only keep regions with sufficient trustable data.
\end{itemize}

Since this is still a ``pseudo ground-truth'' we do not apply LMP on the distillation part $\mathcal{L}_D$ part of the loss but only on the supervised Loss $\mathcal{L}_S$

 \paragraph{Ablations on distillation.}
 \label{sec:distillation_ablation}
 Here we show ablation results for our distillation approach. 
 We compare the results in Tab.~\ref{tab:distillation} after standard pretraining on Flying Chairs and Flying Things 3D to a finetuning on KITTI with and without distillation.
 To gain insights on overfitting and generalization, we provide results on the training datasets as well as cross validation scores on different datasets.
 For completeness we also provide finetuning results of our retrained initial baseline, which uses IPABN with leakyRelu but none of the other improvements.
 The baseline uses the same training schedule, but the original 2k finetuning iterations on Kitti instead of early stopping, as it converges slower since it doesn't use gradient stopping.
 
 It can be seen that standard finetuning leads to high gains on the training datasets, especially \kitF but also drastically reduces performance on the other non-finetuning datasets (which is not surprising).
 Compared to the baseline, the improved model already mitigates this reduction in generalization to some extent, while performing better on the target dataset. %
 Using our proposed distillation approach further improves this generalization to unseen datasets. 
 Interestingly, it even leads to a small improvement on the training dataset itself.
 Since we drastically filter the ``pseudo ground-truth'' it could mean that, this  additional information acts like additional augmentation that benefits the finetuning. 
 
\begin{table*}
	\centering
		\caption{Ablation on Distillation for KITTI finetuning. Comparing pretraining vs.  finetuning (\textsc{FT}) vs. finetuining using distillation (\textsc{Dist}).
		Non baseline models use CV-range $\pm8$ and all proposed improvements  
		(Highlighting \tbf{best} and \und{second-best} results).}
	\label{tab:kitti_distillation}
	\resizebox{0.99\textwidth}{!}{
		\begin{tabular}{ccc|ccc|ccc|cccccc}
			\toprule
			&&&   \multicolumn{3}{c|}{\kitT}    &   \multicolumn{3}{c|}{\kitF}    & \multicolumn{2}{c}{\fc}  &  \multicolumn{2}{c}{\ft}  & \multicolumn{2}{c}{\sintel final} 
			\\
			&&&   \epe [1]   &    \pol [\%]   &  \pol [\%]&   \epe [1]   &    \pol [\%]  &  \pol [\%]  &  \epe [1]  &  \pol [\%]  &  \epe [1]  &  \pol [\%]   &  \epe [1]  &   \pol [\%]   
			\\   
			\textsc{baseline} & \textsc{Dist} & \textsc{FT}     
			&   \multicolumn{2}{c}{train} &test    &   \multicolumn{2}{c}{train}& test    & \multicolumn{2}{c}{\fc}  &  \multicolumn{2}{c}{\ft}  & \multicolumn{2}{c}{\sintel} 
		    \\
		    \midrule
		    % Before finetuning 
			\xmark&\xmark&\xmark
			&    2.37    &    8.65 \%   &- &     7.09    &    18.93 \%  &-  & \tbf{2.31} & \tbf{8.6\%} & \tbf{5.77}& \tbf{11.5\%} &   \tbf{ 4.68}    &  \tbf{11.4\%}  \\       
			\midrule
			% Basline retrained with iPABN for 2000 iterations (like in HD3)
			\cmark&\xmark&\cmark  % 11,23	24,0%	61,34	47,7%	26,77	23,7%
			&    (0.85)    &    (2.35 \%)   &- &     (1.38)    &    (4.41 \%)  &-  & 11.23 & \und{24.0}\% & 61.34 & 47.7\% &   26.77    &  23.7\%  \\       
			\xmark&\xmark&\cmark
			&    \und{(0.77)}    &    \und{(1.91 \%)}   & -&    \und{(1.18)}    &   \und{ (3.58 \%)}  &- &    9.77    &   36.8\%    &   27.92    &    46.2\%    &   \und{ 9.36 }   &     21.5\%     \\
			\rowcolor{mapillarygreen}
			\xmark&\cmark&\cmark
			& \tbf{(0.76)} &    \tbf{(1.84 \%)}   & \tbf{2.25} & \tbf{(1.14)} & \tbf{(3.28 \%)}& \tbf{6.35} &   \und{6.29}    &   24.7\%   &   \und{27.57}    &    \und{38.0\%}    &    9.39    &     \und{18.9\% }    \\
			\bottomrule
		\end{tabular}
	}
	\label{tab:distillation}
\end{table*}

\vspace{-0.75cm}
\subsection{Extending search range}
\label{sec:ablation_extCV}
Here we investigate the impact of extended search ranges in various configurations of our model.
The base configuration always uses gradient stopping and SAD for the cost volume construction and is trained on Flying Chairs2 in forward and backward direction. 
Results for Flying Things 3D are presented to give an insight on generalization on the closest related dataset.
We compare Cost Volume Sampling vs. Cost Volume Warping, both in combination with LMP.

What can be seen from the data, is that LMP clearly helps to improve \epe and \pol metrics in all cases.
What can also be seen, is that in general extending the search range leads to better performance. However, for the combination of Sampling and LMP there is a gain of $0.06$ in \epe when going from a range of  $\pm$4 to $\pm$8, while for the same settings without LMP the total improvement is just $0.01$ and with warping it is $0.04$. We do not experience significant gains that warrant a search range extension of more than $\pm$8 (see Tab.~\ref{tab:ExtSearchCV}). We therefore recommend a version with $\pm8$, Cost Volume Sampling and LMP as it leads to satisfactory performance.

\begin{table*}[h]
   \centering     
   \caption{Extending the cost-volume range leads to lower errors, especially when combined with LMP and Sampling. Model was trained on \fc.} 
     \label{tab:Ablation_CostVol}
     \resizebox{0.6\textwidth}{!}{
 \begin{tabular}{ccc|cccc}
 \toprule
 \textsc{Warp}/ & \textsc{Range} & \textsc{LMP} & \multicolumn{2}{c}{\fc} & \multicolumn{2}{c}{\ft} \\
 Sample & +/- &   & \epe [1] & \pol [\%] & \epe [1] & \pol [\%] \\
 \midrule
     W &         4 &      &                    1.20 &                        6.18\% &                    14.84 &                        25.25\% \\
     W &         8 &      &                    1.16 &                        5.96\% &                    14.20 &                        24.18\% \\
     S &         4 &      &                    1.18 &                        6.15\% &                    15.14 &                        25.00\% \\
     S &         6 &      &                    1.16 &                        6.02\% &                    13.92 &                        24.13\% \\
     S &         8 &      &                    1.17 &                        5.97\% &                    13.46 &                        23.52\% \\
     S &        10 &      &                    1.15 &                        5.92\% &                    15.06 &                        23.44\% \\
     \hline
     W &         4 &    \cmark &                    1.17 &                        6.00\% &                    14.12 &                        23.47\% \\
     W &         8 &    \cmark &                  \und{1.13} &                        5.81\% &                    13.49 &                        23.07\% \\
     S &         4 &    \cmark &                    1.17 &                        5.97\% &                    14.46 &                        23.00\% \\
     S &         6 &    \cmark &                    1.14 &                        5.86\% &                    13.41 &                        23.05\% \\
     S &         8 &    \cmark &                   \tbf{1.11} &                \tbf{5.76\%} &                \und{12.97} &                      \und{22.41\%} \\
     S &        10 &    \cmark &                   \tbf{1.11} &               \und{ 5.78\%} &              \tbf{12.78} &                        \tbf{22.10\%} \\
 % \rowcolor{mapillarygreen}
 \bottomrule
 \end{tabular}
 }
 \label{tab:ExtSearchCV}
%  \end{adjustbox}
 \end{table*}

\subsection{Histogram of Errors}
\begin{figure}[tbhp!]
    \centering
    \includegraphics[width=0.98\columnwidth]{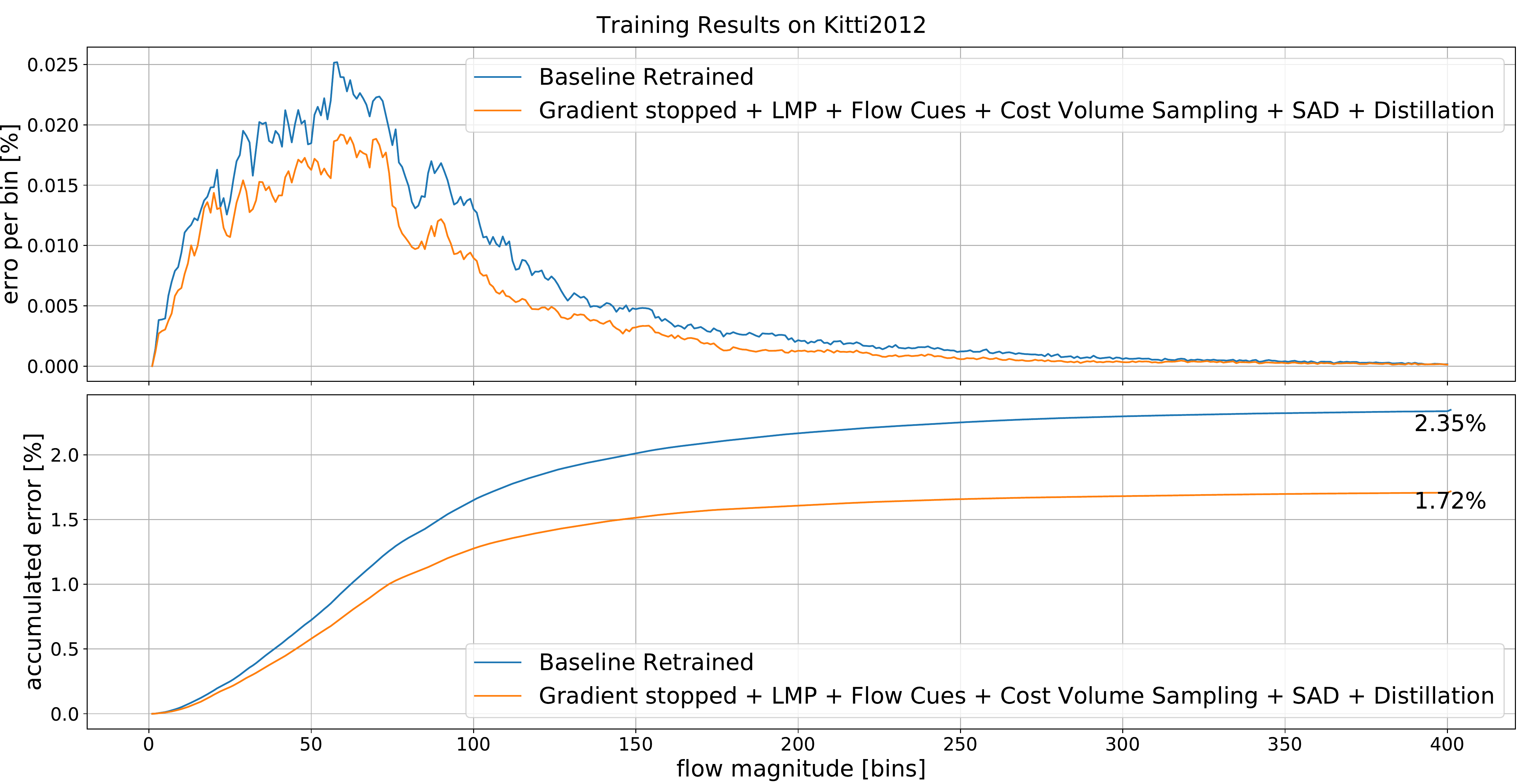}
    \caption{Histogram of errors on the training data of KITTI 2012. The errors are grouped in bins according to the ground truth flow magnitude on which they occurred. Adding all our contributions consistently improves in all areas. }
    \vspace{-1em}
    \label{fig:Hist_errors_2012}
\end{figure}

\begin{figure}[bthp!]
    \centering
    \includegraphics[width=0.98\columnwidth]{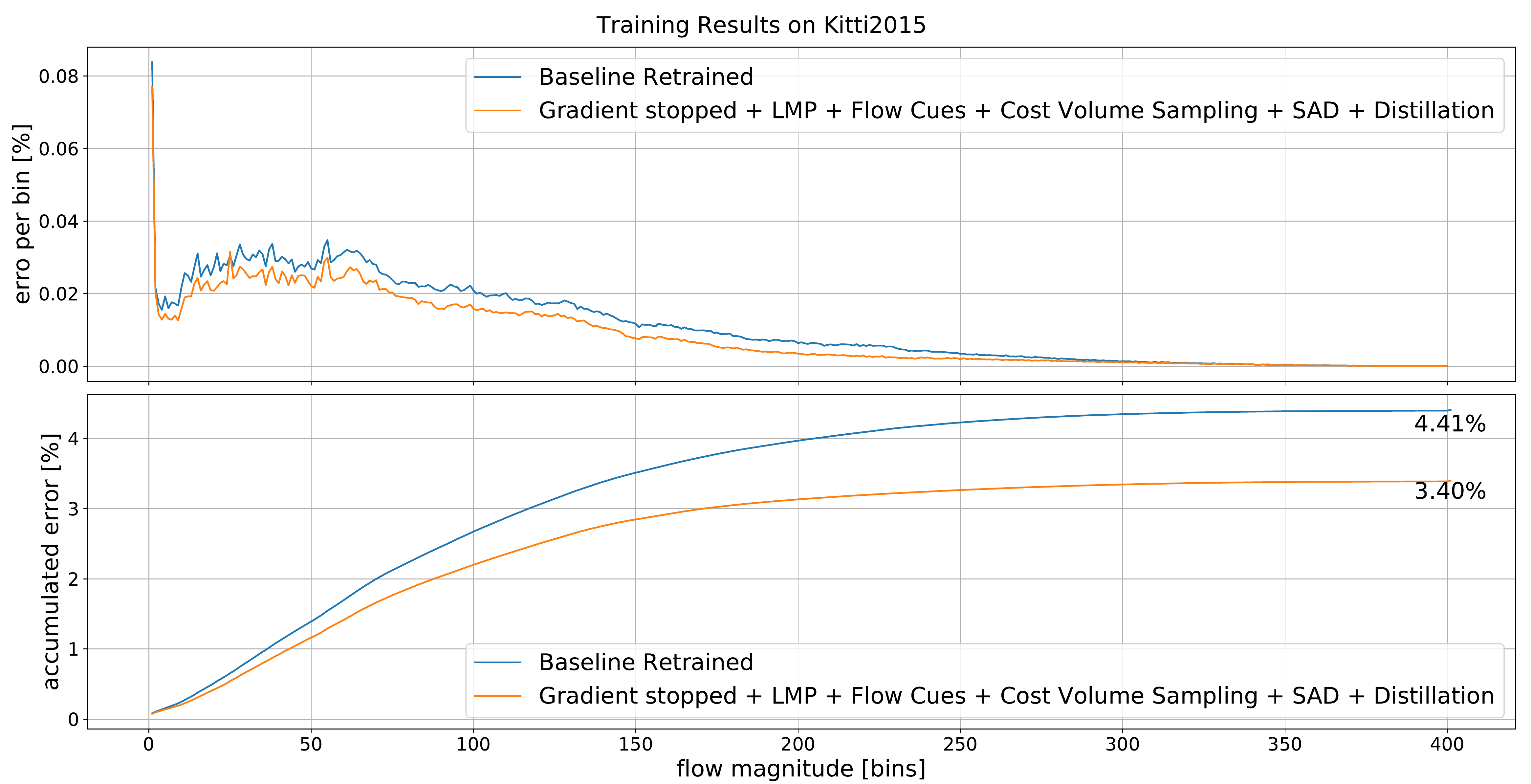}
    \caption{Histogram of errors on the training data of KITTI 2015. The errors are grouped in bins according to the ground truth flow magnitude on which they occurred. Adding all our contributions consistently improves in all areas. }
    \vspace{-1em}
    \label{fig:Hist_errors_2015}
\end{figure}

Fig.~\ref{fig:Hist_errors_2012} and Fig.~\ref{fig:Hist_errors_2015} show the gains made over the KITTI training sequences as achieved with our submitted model that used all proposed improvements and CV-range of $\pm$4. 
The gains are made visible in form of histograms, where the ground truth flow magnitude is used for the binning. As can be seen, our improvements are not limited to a single range of flow magnitudes but affect the whole spectrum of flow vectors.
At this point we want to remind the reader, that adding our contributions hardly changes the number of learnable parameters (e.g. $\approx +1\%$ for HD$^3$) in the network. The gains therefore result from using the provided parameters more effectively. % via our flow elements.

\subsection{Qualitative Comparison of Training Convergence}
\begin{figure}[h!]
	\centering
	\includegraphics[width=\columnwidth]{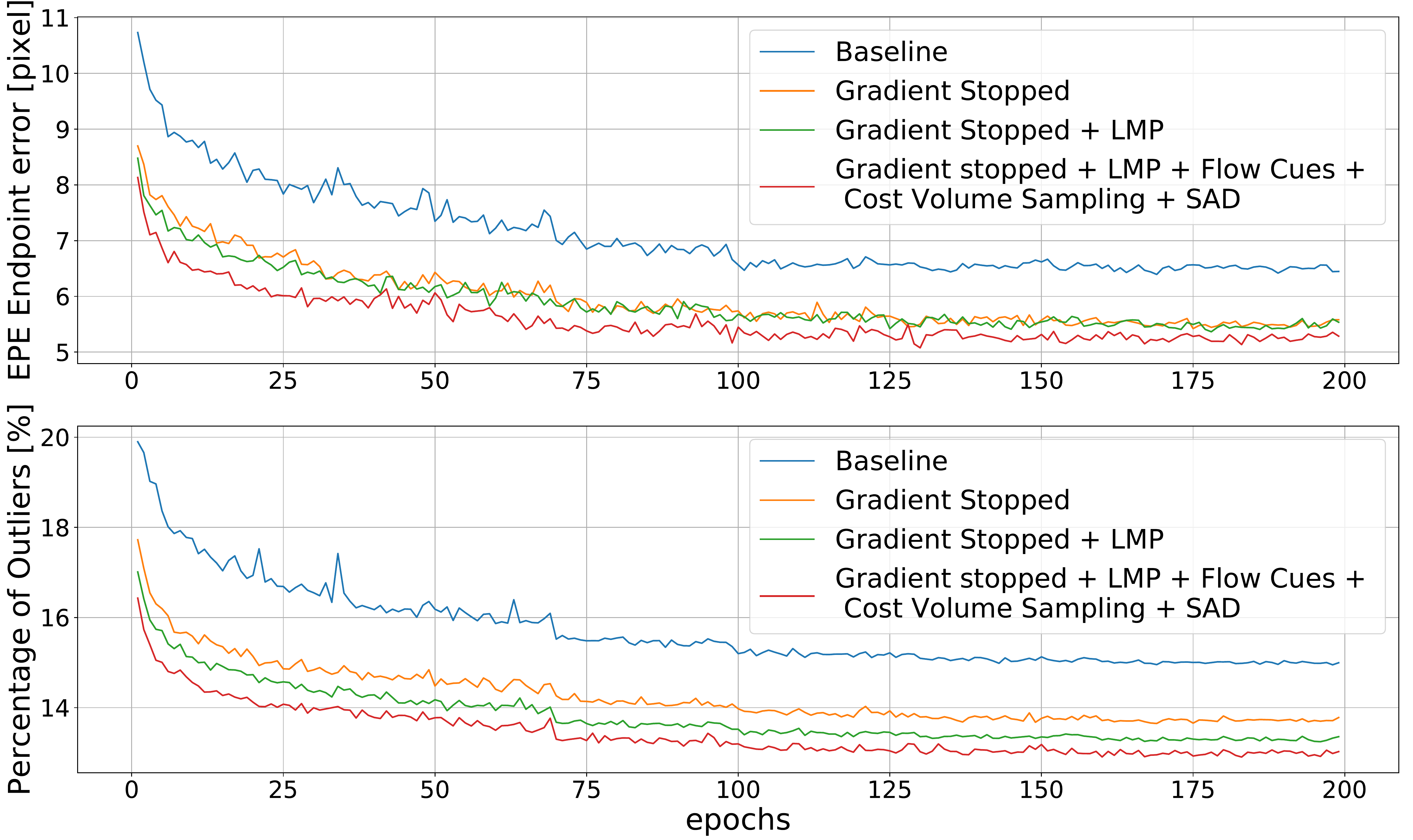}
	\caption{Qualitative comparison of training curves on Things 3D pre-training for optical flow with and HD$^3$ type model (CVr$\pm4$).
		Large drop from Baseline to Gradient Stopped version on EPE and percentage of outliers (\pol). LMP improves mainly on \pol; adding all our remaining contributions gives additional boost on EPE and \pol.
	}
	\vspace{-1em}
	\label{fig:LossCurve_QualitativeComparisonThings3D}
\end{figure}

Fig.~\ref{fig:LossCurve_QualitativeComparisonThings3D} shows exemplary validation curves of an HD$^3$ type model during the Flying Things 3D pre-training.
This is the last part of the pre-training stage before finetuning on KITTI or \sintel. 
These comparisons are qualitatively only, as they were conducted on center crops of the forward flow only, to keep extra computation effort during training low.
We evaluate on the same validation split provided by the original HD$^3$ codebase.

The validation curves in Fig.~\ref{fig:LossCurve_QualitativeComparisonThings3D}  illustrate the overall behavior that we observed on the different datasets and models, when adding our different contributions. 
When adding \detach to the baseline there is a significant drop in both \epe and \pol. Adding \textit{Loss Max Pooling} (\textsc{LMP}) on top mostly affects the \pol by focusing on the remaining difficult examples.
Adding our remaining contributions (Data Distillation is only applied on KITTI) leads to an additional boost in performance on both \epe and \pol.

\subsection{Qualitative Comparisons of Training Results on KITTI}
\begin{figure*}[p]
    \centering
    \includegraphics[width=\linewidth]{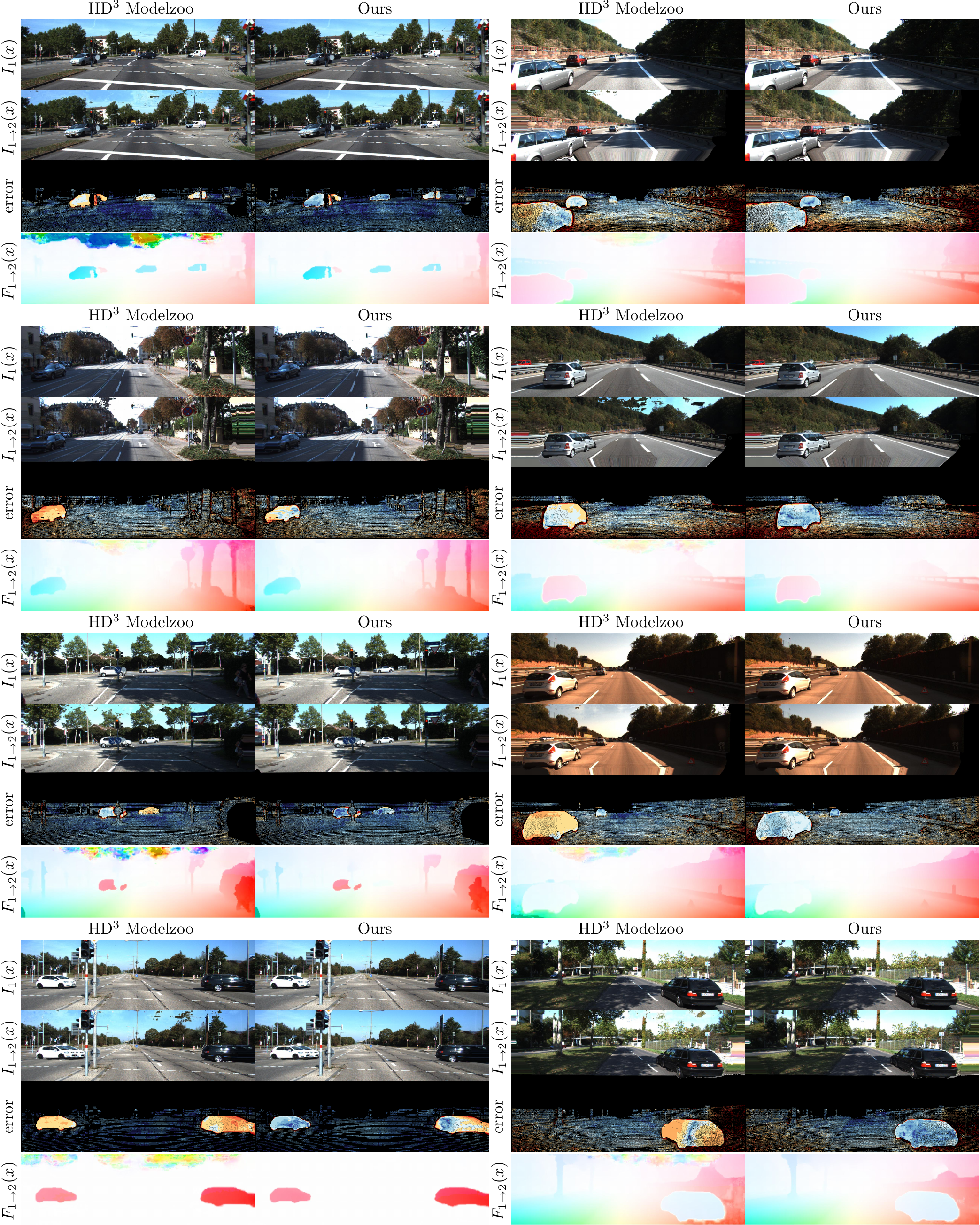}
    \caption{Comparisons on the KITTI 2015 training set. Theirs = HD$^3$ baseline, HD$^3$ with our modifications and contributions and a CV-range of $\pm$4.} 
    \label{fig:ComparisonsKitti2015}
\end{figure*}

In this section various qualitative results on the KITTI training images will be shown.
Fig.~\ref{fig:ComparisonsKitti2015} shows comparisons between the baseline model as taken from the HD$^3$  modelzoo and our best model that uses all our contributions. 
What can be seen especially well in the error plots, is that our model improves a lot on the moving cars.
Furthermore, it improves on fine details, which can e.g. be seen e.g. at the guard rails, where it manages to keep sharper edges and a more homogeneous background.
At the same time, it does not suffer from the artifacts present in the top region of the baseline model. 
The figures are best viewed in high-resolution on a PC.

\subsection{Results on MPI-Sintel}
We outperform the state-of-the-art on the challenging MPI-Sintel Dataset.
Fig. \ref{fig:SintelTest_results} shows the \emph{Results and Rankings} for MPI-Sintel test results at the time of submission to the server. For more details please refer to the main paper.

\begin{figure*}[p]    
	\centering
    \includegraphics[width=1.04\linewidth]{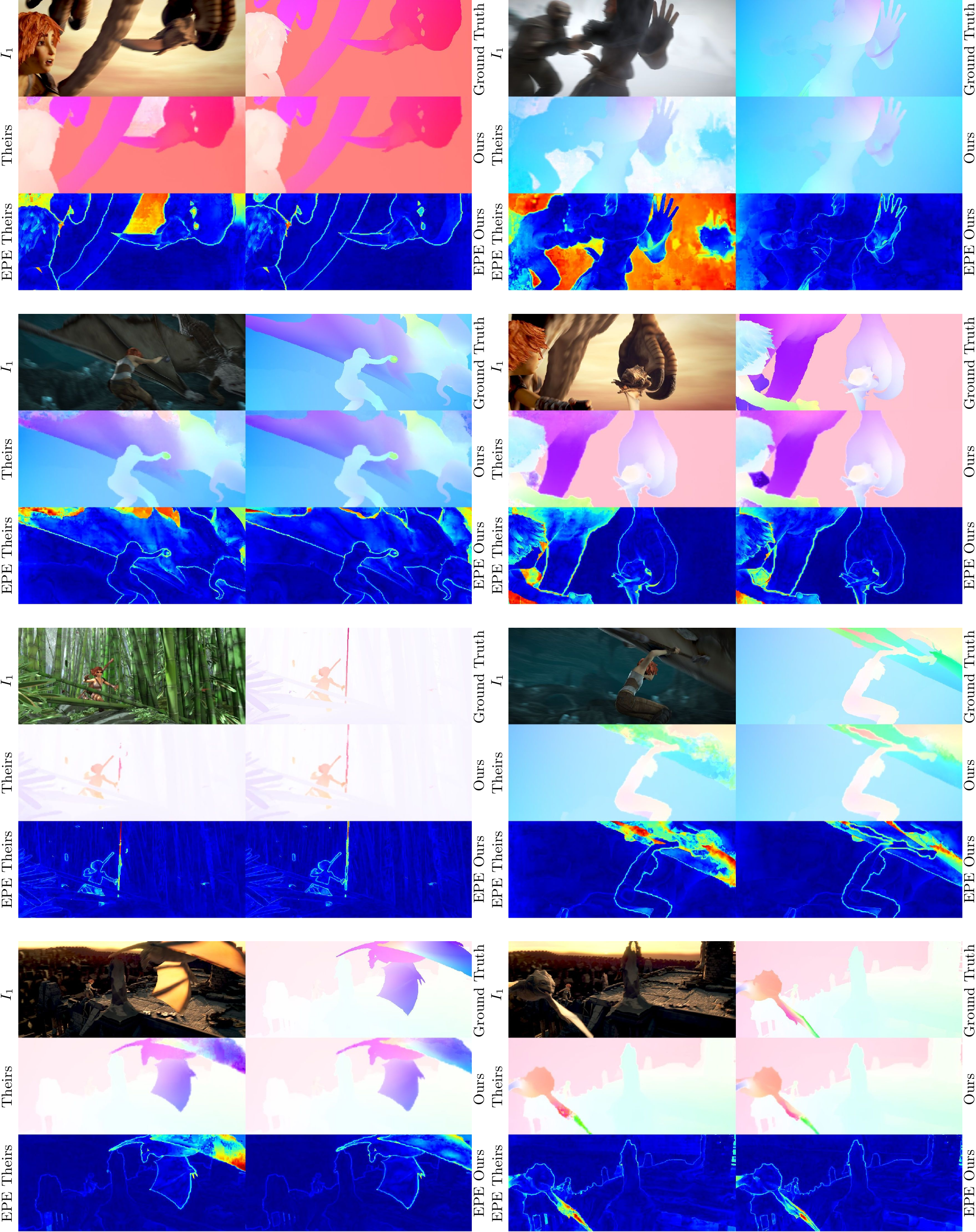}
    \caption{Comparisons on the MPI-Sintel training set. Theirs = HD$^3$ baseline model from the modelzoo. Ours = Adding our contributions on top (Except for Data Distillation since Sintel has dense GT) and a CV-range of $\pm$4. } 
    \label{fig:Comparison_Sintel}
\end{figure*}

\begin{figure*}
    \centering
    \includegraphics[width=\textwidth]{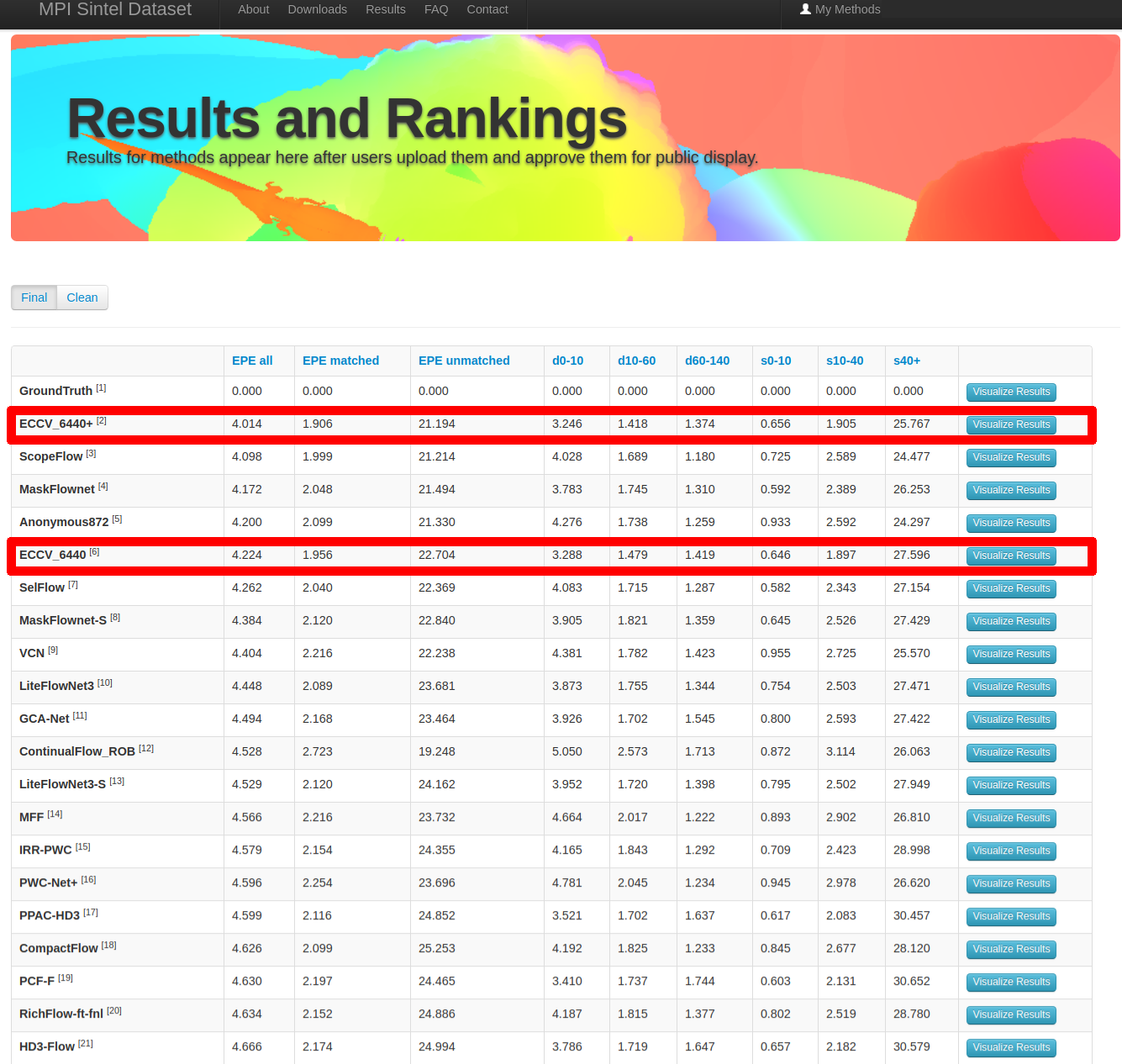}
    \caption{MPI-Sintel \emph{Results and Rankings} - our method improves upon the state of the art. Screenshot taken on March 12, 2020. Short names have been updated after publication to also show IOFPL on benchmark server. } 
    %Screenshot taken at 10th November 2019
    \label{fig:SintelTest_results}
\end{figure*}

Fig.~\ref{fig:Comparison_Sintel} shows the comparison of a HD$^3$ baseline model and our improved baseline trained on the MPI-Sintel training sequence. As can be seen our improved model allows to preserve more fine details like the stick in the bamboo scene or the pike. Also, it seems to be better at detecting and correcting hardly connected moving backgrounds that seem to cause problems for the modelzoo baseline.

\subsection{Sidenote: Sampling vs. Warping -- HD$^3$'s D2V and V2D Operations.}
One of the key innovations in the HD$^3$~\cite{HD3Flow_yin2019hd3}, was the introduction of the D2V and V2D operations that allow to transform match densities into vectors and vice versa.
This operation is used for absolute and residual flows and implicitly assumes an equidistant fixed grid spacing for the flow.
However, this assumption is actually not always valid since the warping operation can deform the space over which the search window operates in the warped image $I_{2\to1}$. I.e. a movement of a single pixel in the search window in $I_{2\to1}(x)$ can move the correspondence to a completely different position in $I_2(y)$ dependent on the flow $F_{2\to1}(x)$  that was used for the warping.

In the case of sampling, the  equidistance of the grid is preserved, since it always uses a single flow vector $F_{2\to1}(x)$ as offset for the entire search window for each individual pixel. Therefore, the spacing of the search window stays equidistant w.r.t. $I_2(y)$ and hence for the D2V and V2D operations.